\documentclass[letterpaper]{article} 

\usepackage[draft]{aaai2026}  
\usepackage{times}  
\usepackage{helvet}  
\usepackage{courier}  
\usepackage[hyphens]{url}  
\usepackage{graphicx} 
\usepackage{natbib}  
\usepackage{caption} 
\usepackage[export]{adjustbox} 
\usepackage[utf8]{inputenc} 
\usepackage[T1]{fontenc}    

\usepackage{amsfonts}       
\usepackage{amsmath}
\usepackage{amssymb}
\usepackage{array} 
\usepackage{booktabs}       
\usepackage{colortbl} 
\usepackage{float}
\usepackage{graphicx}
\usepackage{microtype}      
\usepackage{multicol}
\usepackage{multirow}
\usepackage{natbib}
\usepackage{nicefrac}       
\usepackage{placeins}
\usepackage{subcaption}
\usepackage{xcolor} 

\usepackage{draftwatermark}
\SetWatermarkText{Preprint. \\ Under review.}
\SetWatermarkScale{0.25}
\SetWatermarkColor[gray]{0.85}

\title{PEVLM: Parallel Encoding for Vision-Language Models}

\author{
  Letian Kang\thanks{Corresponding author: \texttt{kangletian@hnu.edu.com}}, Shixian Luo, Yiqiang Li, Yuxin Yin, \\
  Shenxuan Zhou, Xiaoyang Yu, Jin Yang, and Yong Wu \\
  \small\textit{Li Auto Inc.}
}
\date{\today}

\begin{document}

\maketitle

\begin{abstract}
Vision-Language Models (VLMs) have demonstrated strong capabilities in multimodal understanding and generation tasks. However, their application to long video understanding remains hindered by the quadratic complexity of standard attention mechanisms. In this work, we introduce \textbf{PEVLM}, a fine-tuning-free parallel encoding method designed to enhance the prefilling efficiency of VLMs in long video scenarios. PEVLM partitions the input video into context blocks with a shared sink block, while preserving sequential position embeddings to align the attention weight distribution with that of Full-Attention. This design reduces attention complexity from $O((T \times N)^2)$ to $O(T \times N)$ where $T$ is the number of frames and $N$ the number of tokens per frame, without sacrificing accuracy.
Extensive experiments across multiple state-of-the-art models and benchmarks demonstrate that PEVLM consistently outperforms existing parallel encoding approaches, achieving up to \textbf{7.47x} speedup in attention computation and reducing end-to-end latency by \textbf{40\%}. Remarkably, PEVLM not only maintains high accuracy, but in some settings even surpasses Full-Attention performance. Under strict latency constraints, it achieves substantial gains, improving accuracy from \textbf{23.26\%} to \textbf{61.03\%}. These results underscore the effectiveness of PEVLM for low-latency, long-context video understanding, making it a promising solution for real-world applications.
\end{abstract}

\section{Introduction}

In recent years, Vision-Language Models (VLMs) have become a central research focus at the intersection of computer vision and natural language processing. These models have demonstrated impressive performance across a wide range of multimodal understanding and generation tasks \citep{alayrac2022flamingo, li2020oscar, radford2021learning, zhu2023minigpt4}. As their capabilities continue to improve, VLMs are being applied in increasingly complex domains, including robotics \citep{black2024humanoid, cheang2024robot, prasad2024embodied}, autonomous driving \citep{gao2024perception, hu2023drivevlm, wang2024cascaded}, and healthcare \citep{liu2024bmed}. These application scenarios often demand processing longer video inputs, placing significant strain on computational resources.

A key obstacle to applying VLMs to long-video inputs is the quadratic complexity of transformer attention during the prefilling stage \citep{vaswani2017attention, beltagy2020longformer}. This computational burden limits scalability in tasks such as perception, temporal reasoning, and action planning, which are fundamental to domains like robotics, autonomous driving, and healthcare. As a result, optimizing the prefilling phase is critical to enabling efficient long-video processing with VLMs.

To address this issue, a widely adopted technique in large language models (LLMs) is the parallel encoding mechanism \citep{li2024focusllmpreciseunderstandinglong, liu2024star, ma2025blockattentionefficientprefilling, ratner2023parallelcontextwindowslarge, yang2025ape, yen2024longcontextlanguagemodelingparallel, lu2024turboragacceleratingretrievalaugmentedgeneration}. In this approach, the input context is divided into multiple blocks, with each block independently encoded into Key-Value (KV) states. This design prevents tokens in different blocks from attending to one another, thereby reducing the computational complexity from $O(L^2)$ to $O(L)$. Moreover, parallel encoding alleviates the issue of “lost in the middle” by reducing the number of tokens participating in the softmax operation and can even achieve accuracy surpassing that of Full-attention in long-context scenarios \citep{Liu2023LostIT, yang2025ape, 2025softmaxforsharpsize}.

Several state-of-the-art architectural improvements have been proposed to enhance the parallel encoding mechanism. \textbf{Adaptive Parallel Encoding (APE)} improves prefilling efficiency by introducing shared prefixes, an attention temperature parameter, and a scaling factor to better align the distribution of attention weight in parallel encoding with that of sequential encoding. Although these adjustments improve accuracy, they are highly sensitive to length, quantity, and content, making them difficult to tune and deploy in practice \citep{yang2025ape}.
\textbf{Star Attention} extends ring attention by incorporating a parallel encoding strategy, which eliminates communication overhead during the prefilling stage and reduces overall computational complexity \citep{liu2024star}. Another notable distinction of Star Attention is the use of a larger sink size, which is equal to the block size.
\textbf{Block-Attention \& TurboRAG} share the same idea: build on parallel encoding, apply position re-encoding and fint-tune to improve accuracy \citep{guu2020retrieval, lu2024turboragacceleratingretrievalaugmentedgeneration}. However, without fine-tune, the accuracy of both suffers for the absence of a Sink Block. Nevertheless, the idea of preserving the sequential position embeddings for each block, rather than reusing a shared one, offers a valuable and insightful idea.

In addition, there are several widely discussed methods such as Native Sparse Attention \citep{yuan2025nativesparseattentionhardwarealigned} from Deepseek and MoBA \citep{lu2025mobamixtureblockattention}, which adopt similar algorithmic principles and achieve acceleration effects comparable to parallel encoding. However, both approaches involve modifications to the model architecture and thus require retraining from scratch. This poses a challenge for many inference platforms, which typically do not have control over users' model preferences. The issue is particularly pronounced in domains such as autonomous driving, where a reliable models are often continuously trained and iterated over years. Consequently, we focus on finetune-free performance optimization methods to ensure compatibility and ease of deployment.

Given all of these, a key question naturally arises: \textbf{Can existing methods be directly applied to VLMs to accelerate inference and deployment pipelines?}
In Table~\ref{tab:algo_acc}, we evaluate several models with existing parallel encoding methods on LongVideoBench \citep{longvideobench2024}, VideoMME \citep{fu2024video}, EgoSchema \citep{NEURIPS2023_90ce332a} and MVBench \citep{li2024mvbenchcomprehensivemultimodalvideo} benchmarks. \textbf{Although existing parallel encoding methods demonstrate certain effectiveness on shorter videos, we observe varying degrees of accuracy degradation across different models when applied to longer videos.} This performance drop is particularly pronounced in models like Qwen2.5-VL, which are more sensitive to spatial and temporal information in videos—sometimes resulting in completely empty or garbled outputs in long-video scenarios, with accuracy dropping to 0.0\%. To investigate this issue, we analyze the attention weight distributions under parallel encoding and identify a misalignment between Full-Attention and parallel encoding. Existing methods developed for LLMs fail to effectively address this misalignment in VLMs.
Building upon this observation, we conducted a series of studies and experiments and ultimately propose a novel attention mechanism for VLMs, termed \textbf{Parallel Encoding for Vision-Language Models (PEVLM)}. PEVLM is a fine-tuning-free method specifically designed to accelerate the prefilling stage in long-video processing with VLMs.
Our contributions involve:
\begin{itemize}
  \item We systematically analyze the distribution characteristics of attention weights under parallel encoding in VLMs. Our observations reveal a key reason for the misalignment between the attention weight distributions of parallel encoding and Full-Attention: the reuse of position embeddings across blocks leads to the loss of critical video information.
  \item We propose PEVLM to recover the accuracy of parallel encoding by applying three targeted alignment steps: (i) segmenting contexts into blocks by video frames rather than tokens; (ii) using the system promots and the initial frames of the video as Sink Block for all blocks to avoid the duplication of abnormal distribution of initial tokens; (iii) preserving the sequential position embeddings instead of resuing position embeddings across blocks. With these alignment strategies, PEVLM achieves higher accuracy than existing parallel encoding methods, and reaches 99.57\% to 104.80\% of the accuracy of full attention on different models.
  \item PEVLM reduces the computational complexity of the attention mechanism from $O((T \times N)^2)$ to $O(T \times N)$, which leading to a significant improvement in prefilling efficiency.
  (i) For 100k (text\&video) token contexts prefilling, the attention layer achieves a 7.47× speedup, while the end-to-end 2.58x speedup without compromising generation quality. This highlights the practical viability and superiority of PEVLM in cloud deployment scenarios.
  (ii) Under fixed latency constraints, PEVLM increases accuracy from 23.26\% to 61.03\%, showcasing its critical value for latency sensitive applications.
\end{itemize}

\section{Observations}
\label{sec:observations}

\begin{figure*}[h]
    \centering
    
    \begin{subfigure}[t]{0.32\textwidth}
        \centering
        \includegraphics[width=\linewidth]{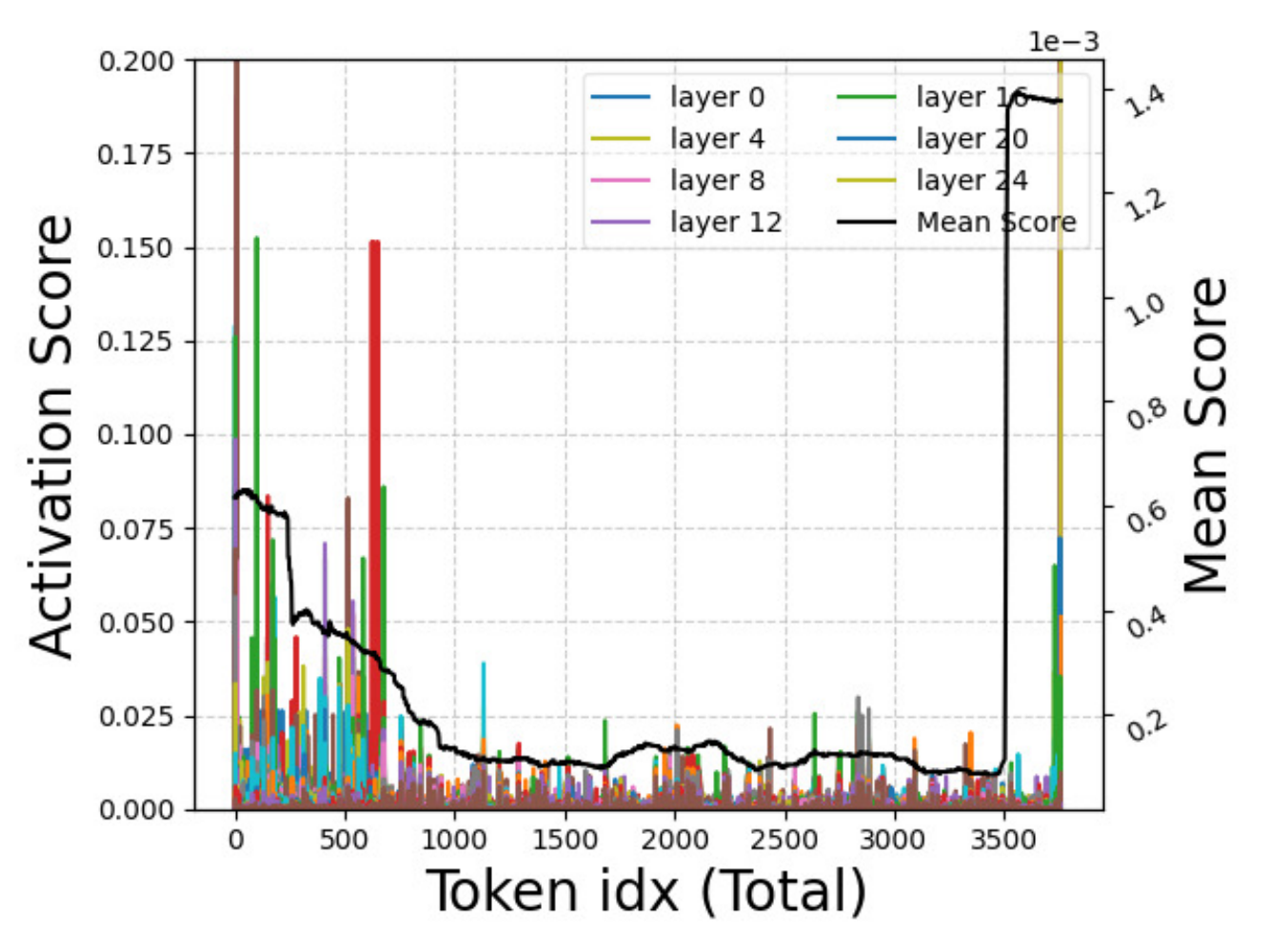}
        \includegraphics[width=\linewidth]{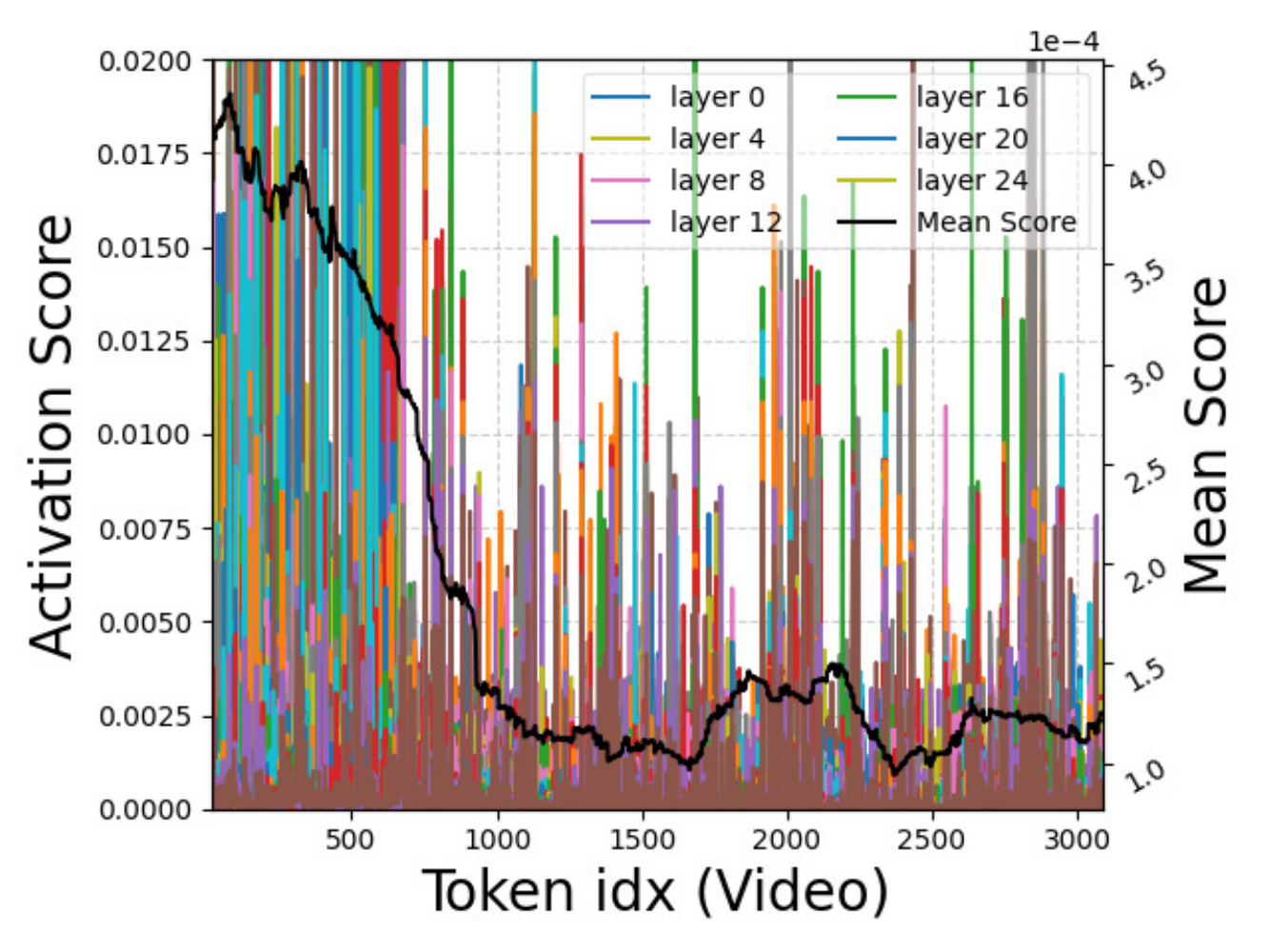}
        \caption*{(a) Full-Attention}
    \end{subfigure}
    \hfill \raisebox{-0.5\height}{\textcolor{lightgray}{\vrule width 0.8pt height 8cm}} \hfill 
    \begin{subfigure}[t]{0.32\textwidth}
        \centering
        \includegraphics[width=\linewidth]{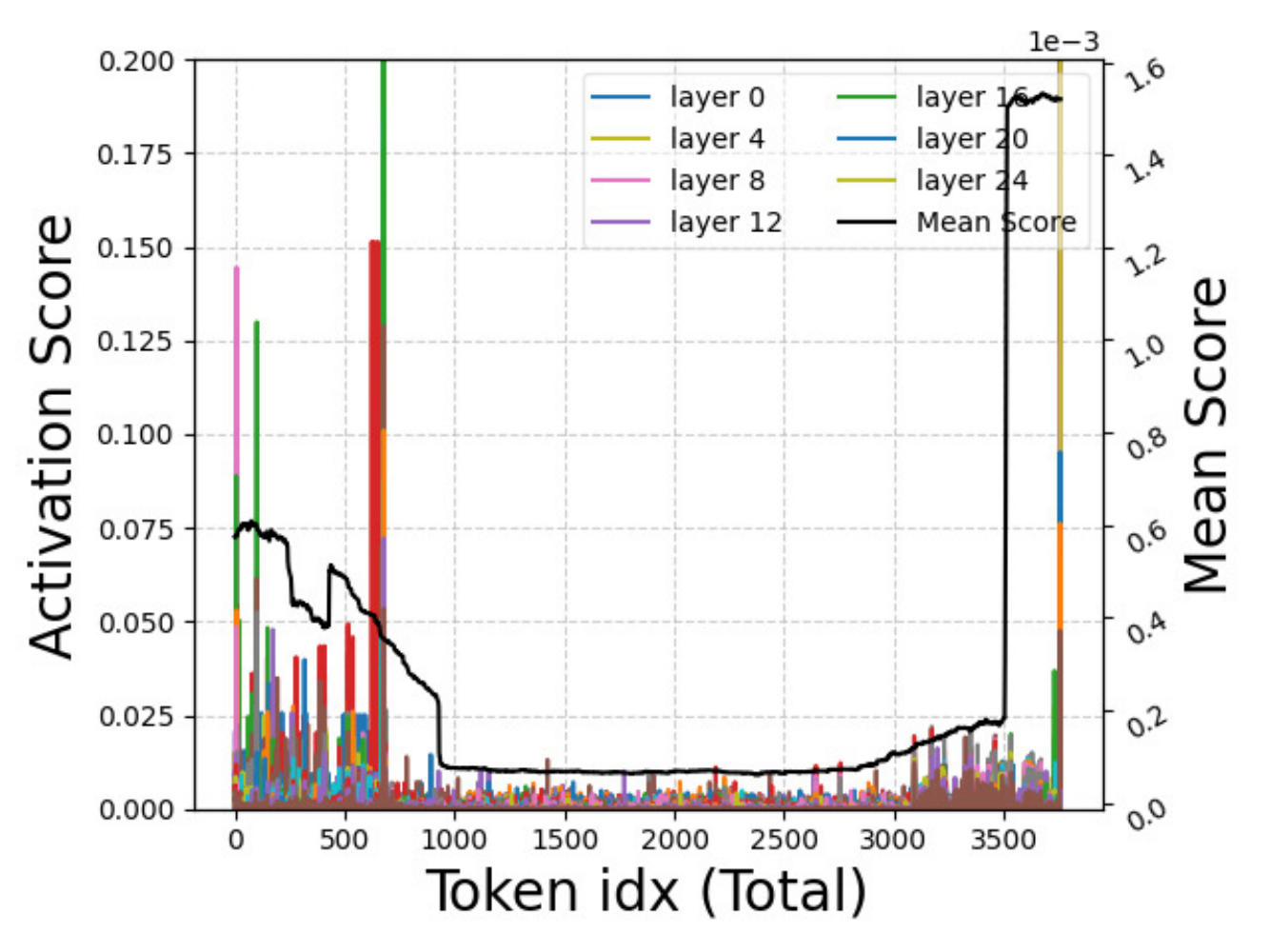}
        \includegraphics[width=\linewidth]{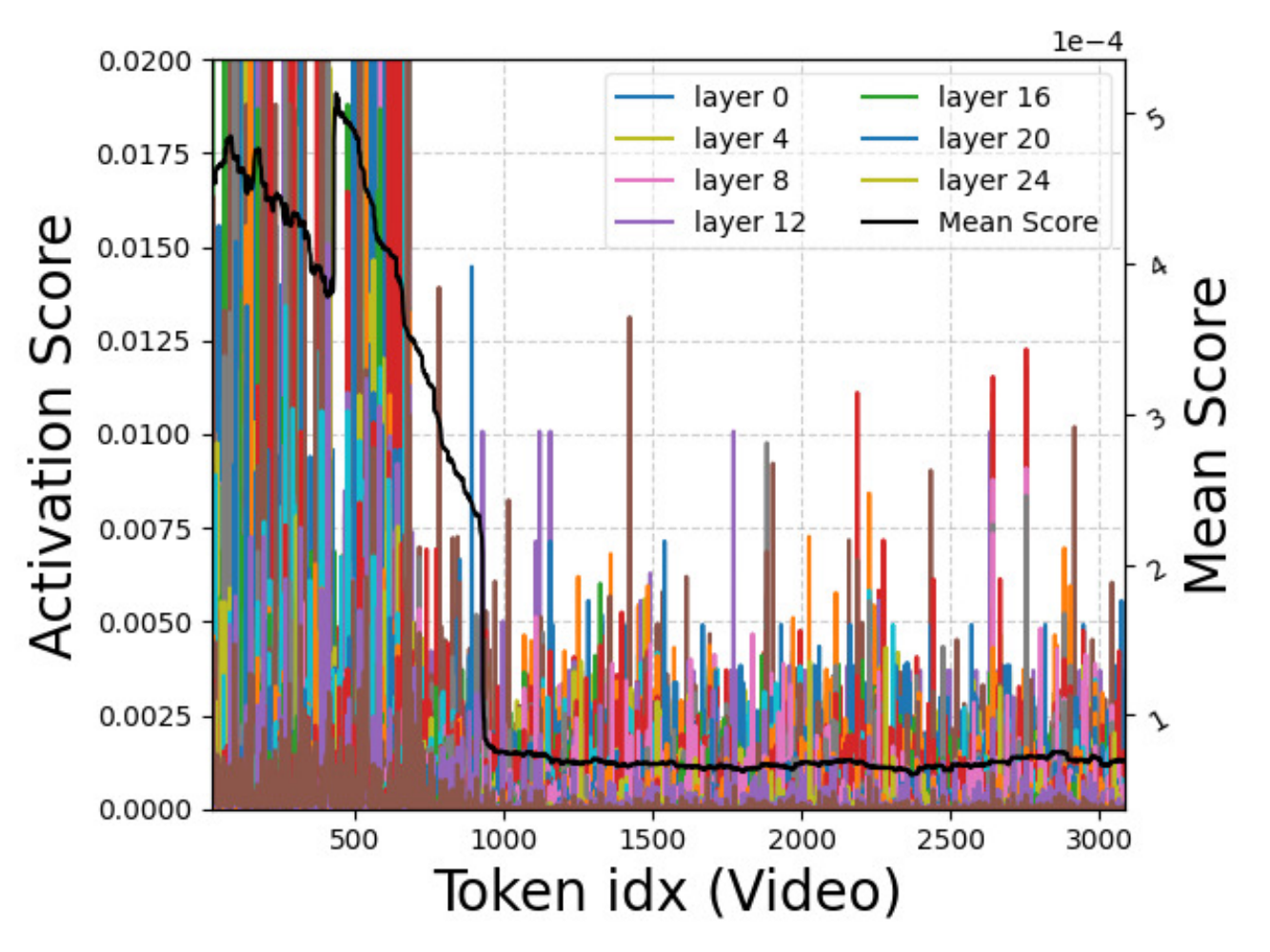}
        \caption*{(b) Parallel Encoding (Reusing Position Embedding)}
    \end{subfigure}
    \hfill \raisebox{-0.5\height}{\textcolor{lightgray}{\vrule width 0.8pt height 8cm}} \hfill 
    \begin{subfigure}[t]{0.32\textwidth}
        \centering
        \includegraphics[width=\linewidth]{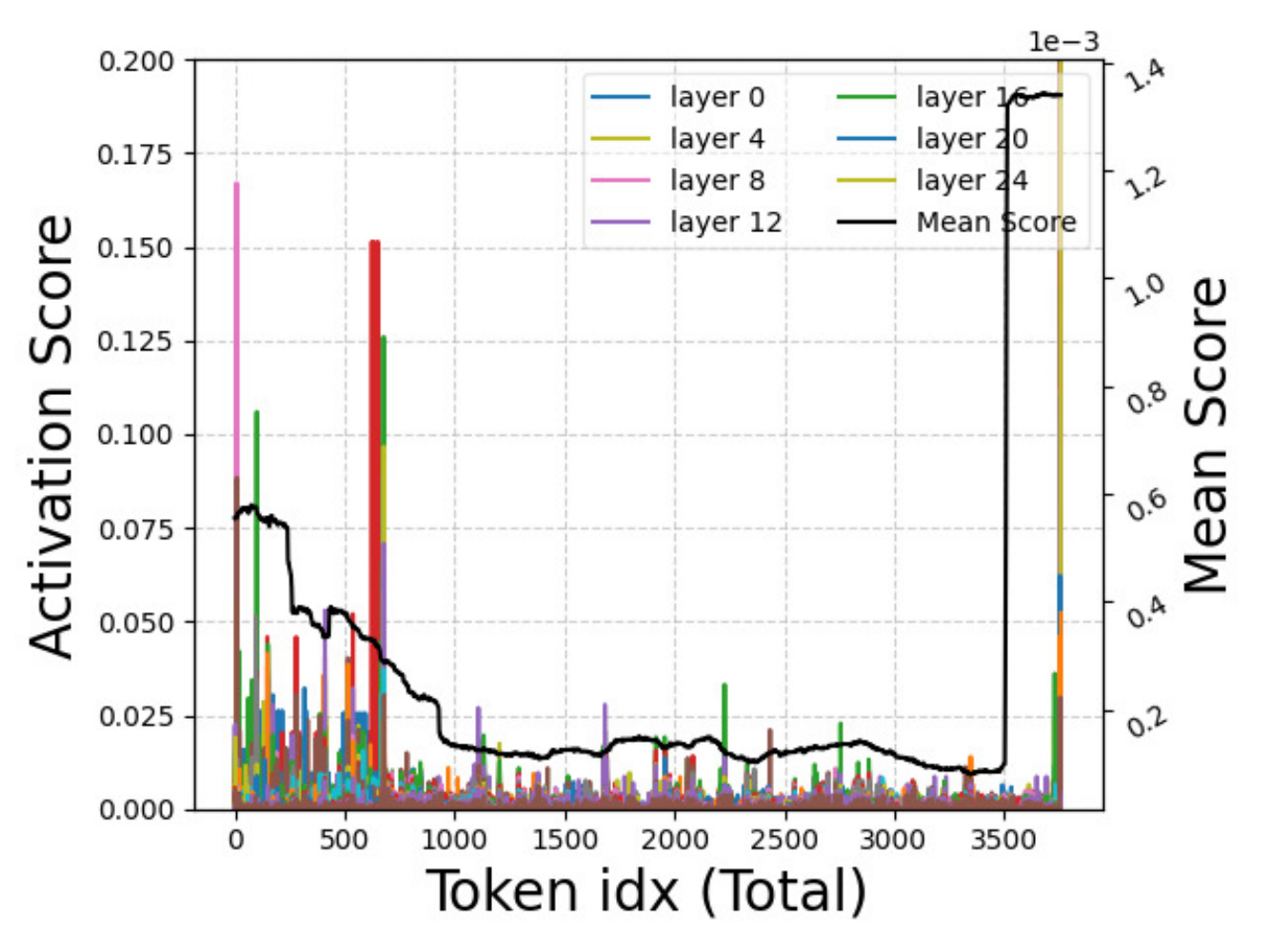}
        \includegraphics[width=\linewidth]{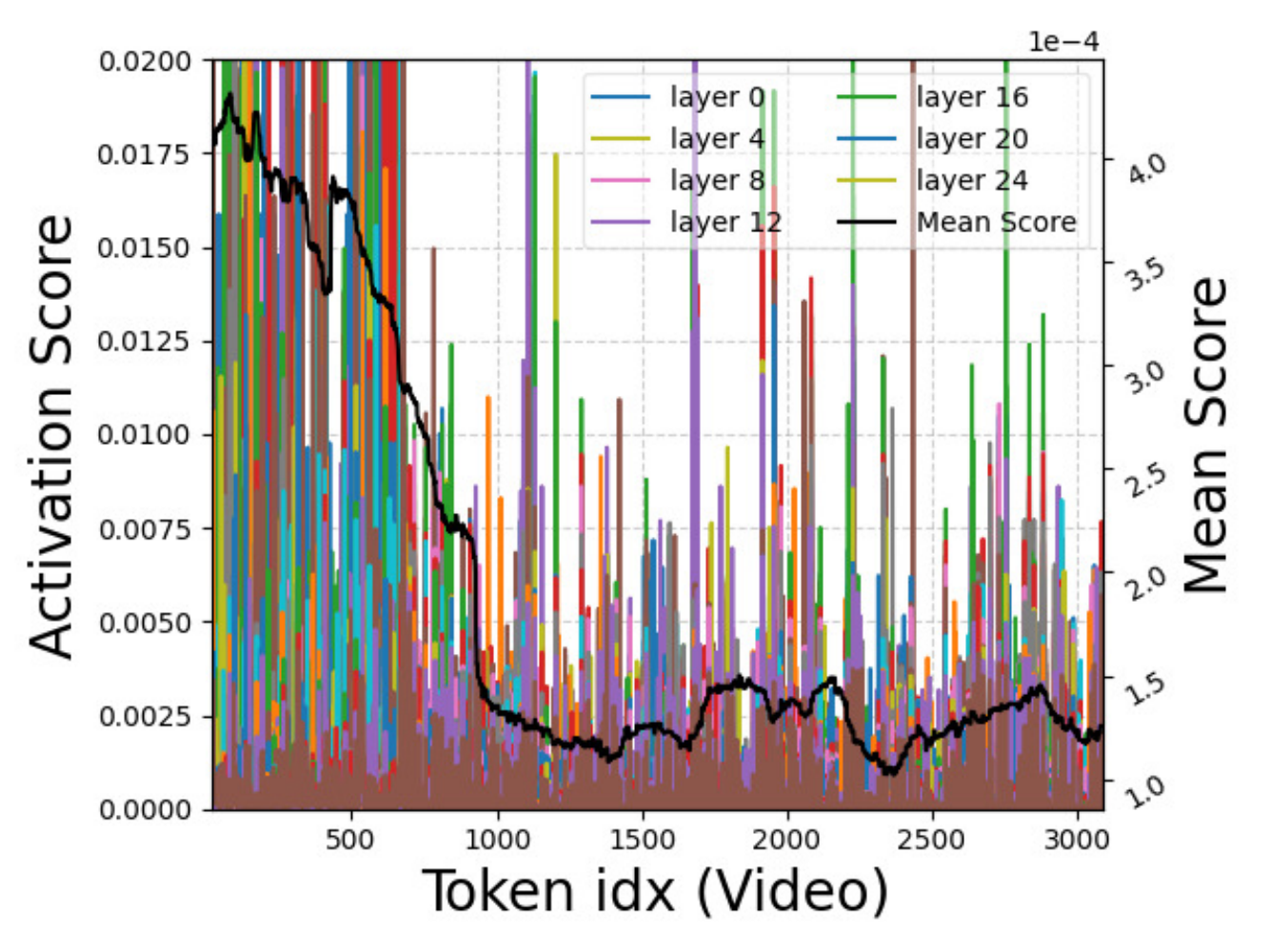}
        \caption*{(c) Parallel Encoding (Sequential Position Embedding)}
    \end{subfigure}
    
    \caption{We collected the distribution of attention weights under different scenarios and computed their moving averages for better observation. The top row shows the attention weights for all tokens, while the bottom row shows those for the visual tokens. (a) Full-Attention: The textual tokens and the initial video frames receive the highest attention weights, whereas the remaining video frames are assigned relatively lower attention weights. (b) Parallel Encoding (Reusing Position Embedding): The overall attention weights distribution resembles Full-Attention, with higher attention weights for the textual tokens and initial video frames. While compared to Full-Attention, the initial video frames receive even higher attention weights, and the remaining video frames exhibit more uniform and lower attention weights. (c) Parallel Encoding (Sequential Position Embedding): The overall attention weights distribution remains similar to (a) and (b). However, the attention weight distribution within the video frames more closely aligns with (a)).}
    \label{fig:attention_weights}
\end{figure*}

To explore the application of parallel encoding in VLMs, we analyzed the potential differences between LLMs and VLMs that may affect accuracy:

(i) \textbf{In VLMs, attention sinks are present not only in the initial system prompt, but also in the early video frames} ~\citep{kang2025toldvisualattentionsink}. To achieve optimal performance, it may be necessary to include enough frames in the sink.

(ii) Existing parallel-encoding-based methods are primarily designed for Retrieval-Augmented Generation (RAG) tasks \citep{ma2025blockattentionefficientprefilling, yang2025ape, lu2024turboragacceleratingretrievalaugmentedgeneration}. In such tasks, reusing the KV cache from previously retrieved passages is crucial for enhancing inference efficiency. Since the order of passages often changes when reusing the KV cache, a common strategy is to apply parallel encoding to generate the KV cache for each passage independently, while sharing position embeddings across passages, as illustrated in Figure~\ref{fig:position_attn}. As a result, position embeddings are typically shared across blocks. However, \textbf{preserving the correct temporal order of video frames is critical for video understanding tasks}. Blindly reusing position embeddings across blocks can disrupt the model's ability to capture temporal dependencies. This disruption is especially severe for models where the temporal component of the position embeddings is explicitly aligned with absolute time \citep{bai2025qwen25vltechnicalreport}.

Therefore, as illustrated in Figure~\ref{fig:attention_weights}, we perform an attention weight analysis for VLMs and make two key findings:

(i) As shown in Figure~\ref{fig:attention_weights}(a), some initial text and visual tokens receive high attention scores, consistent with the observations reported in \citep{mit2023streamingllm, sun2024massiveactivationslargelanguage, kang2025toldvisualattentionsink}. This suggests that it may be necessary to include early video frames when selecting sink tokens.

(ii) As shown in Figure~\ref{fig:attention_weights}(b), compared to Full-Attention, the initial video frames receive even higher attention weights, while the remaining frames exhibit more uniform and lower attention weights. In contrast, when sequential position embeddings are preserved, as illustrated in Figure~\ref{fig:attention_weights}(c), parallel encoding yields an attention score distribution that more closely resembles that of Full-Attention.

These observations yield two key insights for applying parallel encoding to VLMs: (i) the shared sink block in parallel encoding should include early video frames in addition to textual prompts, as they function as important attention sinks; (ii) maintaining the sequential position embeddings of visual tokens is essential for preserving temporal consistency, and reusing position embeddings across blocks could disrupt temporal alignment. These findings underscore the need for a VLM-specific parallel encoding strategy, which we introduce in the next section.

\section{PEVLM}
With the observation in the last section, we design PEVLM to improve computational efficiency. PEVLM enables a seamless shift from Full-Attention to parallel encoding without requiring training while maintaining most of the model's capabilities. Our approach adaptively aligns the distribution of attention weights between Full-Attention and parallel encoding, thereby boosting efficiency and performance.

\subsection{Standard Attention Mechanism}
Before introducing PEVLM, we briefly review key aspects of the standard attention mechanism as a foundation. The computation formula for standard softmax attention is:
\begin{equation}
O=Softmax(\frac{QK^T}{\sqrt{d} })V\ \ \ Q\in \mathbb{R}^{n\times d}\ \ \ K,V\in \mathbb{R}^{m\times d}
\label{eq:fullattn}
\end{equation}
where $Q$ is the query state, and $K$ and $V$ denote the key and value states, respectively.

\textbf{Attention Sink.} The attention sink phenomenon was first identified by StreamingLLM \citep{mit2023streamingllm}, which observed that one or more tokens at the beginning of the input consistently receive higher attention scores. APE \citep{yang2025ape} and Star-Attention \citep{liu2024star} have demonstrated that sharing a common attention sink across parallel encoding blocks significantly enhances the accuracy of the model. Some studies also show that the attention sink phenomenon exists in VLMs but is more distributed among visual tokens \citep{kang2025toldvisualattentionsink}.

\begin{figure*}[th]
    \centering
    \includegraphics[width=\textwidth]{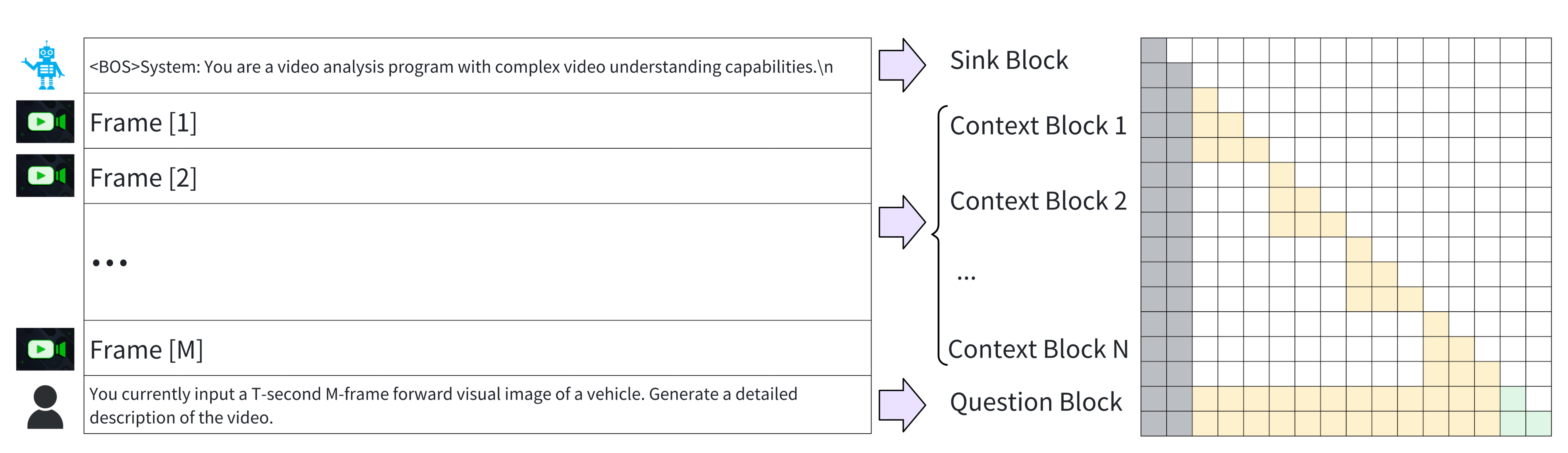}
    \caption{The PEVLM Masks.}
    \label{fig:vlm_PEVLM}
\end{figure*}

\textbf{Position Embedding.} Position embedding provides essential information that distinguishes the position and order amongst tokens in a sequence. For VLMs, contextual positional information is more critical. In particular, for some VLM models like Qwen2.5-VL \citep{bai2025qwen25vltechnicalreport}, their position embeddings also incorporate spatio-temporal information between video frames, making position embeddings increasingly essential for VLMs.

\textbf{Computational Complexity.}
In standard softmax attention mechanisms, the total number of operations in the prefilling phase is given by
\begin{equation}
    OP_{Attn}=2HL^2, \label{eq:ops_attn}
\end{equation}
where $L$ is the context length and $H$ is the hidden size. This leads to a computational complexity of $O(L^2)$.  As $L$ increases, the cost grows quadratically, severely limiting the speed and scalability of the inference. This issue is further exacerbated in VLMs. For a video consisting of $T$ frames, where each frame is represented by $N$ tokens (typically ranging from hundreds to thousands), the total context length becomes $L=T \times N$, resulting in a complexity of $O((T \times N)^2)$. Such scaling renders standard attention mechanisms impractical for long-video understanding tasks.

\subsection{Parallel Encoding}
\subsubsection{Partitioning Strategy}
As illustrated in Figure~\ref{fig:vlm_PEVLM}, our partitioning strategy consists of three core components:

\begin{itemize}
    \item \textbf{Sink Block}: The initial text tokens (e.g., BOS token and system prompts) and the first several video frames are grouped into a dedicated Sink Block.
    \item \textbf{Context Blocks}: The remaining video content is uniformly partitioned into Context Blocks by frames to reduce computational overhead and enable parallelization.
    \item \textbf{Question Block}: The text tokens that follow the video input are left unsegmented as Quesion Block.
\end{itemize}

\subsubsection{Position Encoding}

\begin{figure}[h]
    \centering
    \includegraphics[width=0.45\textwidth]{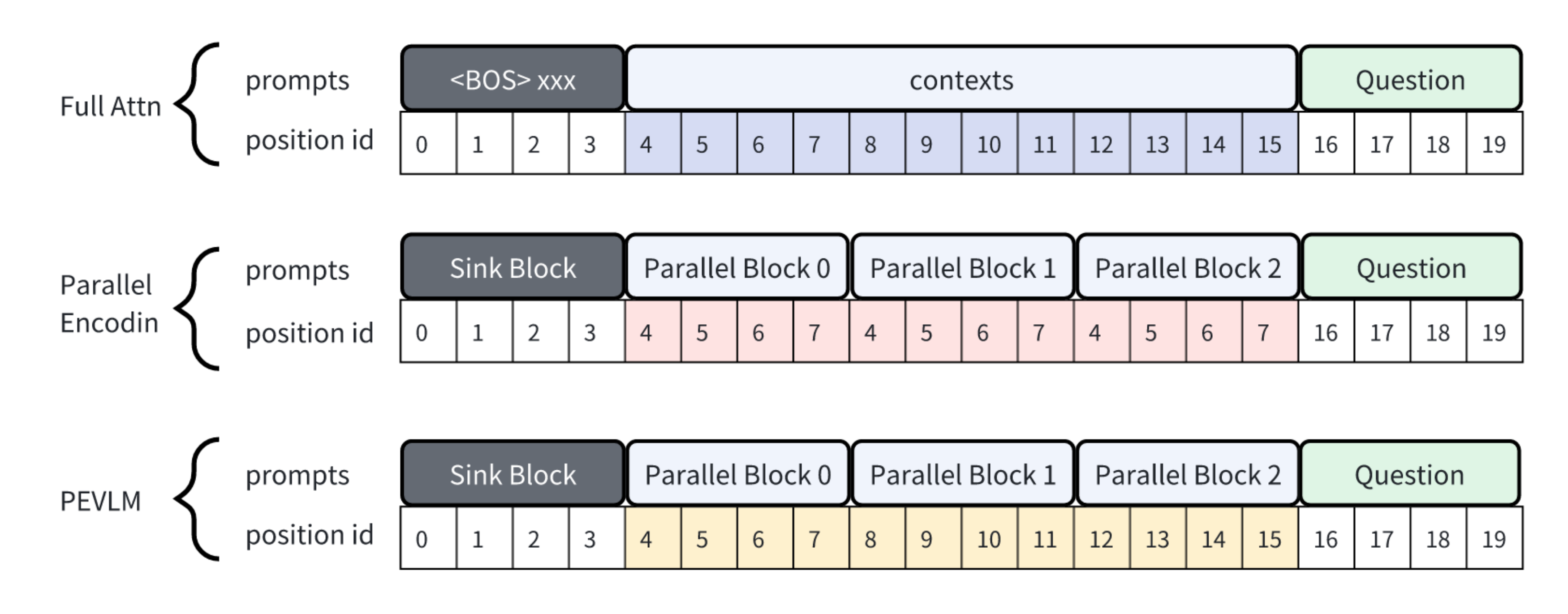}
    \caption{Position Encoding in PEVLM}
    \label{fig:position_attn}
\end{figure}

As described in the \textit{Observations} section, similar to the case in LLMs, applying parallel encoding methods in VLMs suffers from the misalignment of attention weight distributions when compared to Full-Attention. While prior work has attempted to address this issue by introducing additional hyperparameters to realign attention scores \citep{yang2025ape}, such approaches are not ideal for efficient deployment. To address this, PEVLM draws inspiration from Block Attention \citep{ma2025blockattentionefficientprefilling} and TurboRAG \citep{lu2024turboragacceleratingretrievalaugmentedgeneration}, and retains sequential position embeddings, as shown in Figure~\ref{fig:position_attn}.
In contrast to these methods, which dynamically adjust positional embeddings, often to accommodate context reuse through sequence reordering, PEVLM forgoes such updates. Since video frames are processed in their original temporal order, PEVLM directly applies the sequential position embedding to each visual token within context blocks, without modifying their positions after query concatenation.

By preserving sequential position embeddings, PEVLM maintains attention weight distributions that closely resemble those of Full Attention. However, unlike some parallel encoding methods that aim to extend context capacity, this design choice also constrains PEVLM's ability to scale to longer video sequences. For instance, LongVILA-7B-256f is typically trained on sequences up to 256 frames; attempting to process longer inputs may lead to performance degradation. Although PEVLM achieves significant acceleration of inference, it is similarly limited to 256-frame inputs. Nonetheless, this trade-off is acceptable as our primary goal is to enhance inference efficiency rather than to expand context length.

\begin{table*}[h]
    \centering
    \caption{Performance ($\uparrow$) of different models and different methods on video understanding tasks evaluated at tokens from 26k to 100k.}
    \label{tab:algo_acc}
    \begin{tabular}{l l c c c c c c c c c c c c}
        \toprule
        \multirow{2}{*}{Model} & \multirow{2}{*}{Method} & MVBench & EgoSchema  &  \multicolumn{2}{c}{VideoMME} &  \multicolumn{2}{c}{LongVideoBench (Val)} & \multirow{2}{*}{Avg.} \\
        & & $17s\ avg.$ & $3mins\ avg.$ & $<2mins$ & $2mins$-$1h$ & $<1min$ & $1min$-$1h$\\
        \midrule
         Qwen2.5-VL- & Full Attn & \textbf{68.47}\% & 57.00\% & \textbf{74.56}\% & 56.72\% & 72.85\% & 55.33\% &  64.16\%\\
         7B-Instruct & Block Attn &  66.03\% & 47.40\% & 65.22\% & 50.39\% & 70.64\% &47.85\% & 57.92\%\\
         \textit{Frames: 768} & APE(T=1.0) &  66.03\% & 57.20\% & 50.56\% & 1.06\% & 54.02\% & 3.38\%& 38.71\%\\
         \textit{Tokens: 100k} & Star Attn & 60.19\% & 19.80\% & 3.00\% & 0.00\% & 49.03\% & 0.00\% & 22.00\%\\
         & PEVLM & 68.44\% & \textbf{61.00}\% & 74.33\% & \textbf{58.83}\% & \textbf{74.52}\% & \textbf{57.58}\% & \textbf{65.78\%}\\
        \midrule
         LongVILA- & Full Attn &  61.92\% & 57.00\% & 66.33\% & 49.39\% & 65.10\% & 47.44\%& 57.86\%\\
         7B-256f & Block Attn &  57.81\% & 59.00\% & 64.89\% & 51.44\% & 61.77\% &48.05\% & 57.16\%\\
         \textit{Frames: 256} & APE(T=1.0) & 60.00\% & 60.60\% & 68.56\% & 51.61\% & 63.71\% & 46.93\%& 58.57\%\\
         \textit{Tokens: 66k} & Star Attn& 63.39\% & \textbf{61.60}\% & \textbf{70.11}\% & 54.17\% & 63.43\% &  47.44\%& 60.02\%\\
         & PEVLM & \textbf{63.53}\% & 61.40\% & 69.44\% & \textbf{54.33}\% & \textbf{65.93}\% & \textbf{49.18}\%& \textbf{60.64}\%\\
        \midrule
         LLaVA-Video- & Full Attn & \textbf{60.67}\% & \textbf{58.20}\% & \textbf{75.33}\% & \textbf{59.99}\% & \textbf{72.02}\% &  \textbf{55.53}\%& \textbf{63.62\%}\\
         7B-Qwen2 & Block Attn & 56.86\% & 51.60\% & 67.56\% & 55.50\% & 63.99\% & 52.36\%& 57.98\%\\
         \textit{Frames: 128} & APE(T=1.0) & 58.00\% & 57.60\% & 72.22\% & 58.19\% & 67.87\% & 54.61\%& 61.42\%\\
         \textit{Tokens: 26k} & Star Attn & 59.31\% & 57.80\% & 72.89\% & 58.71\% & 68.98\% & 54.92\%& 62.10\%\\
         & PEVLM & 60.00\% & \textbf{58.20}\% & 74.33\% & \textbf{59.99}\% & \textbf{72.02}\% & \textbf{55.53}\%& 63.35\%\\
        \bottomrule
    \end{tabular}
\end{table*}

\subsubsection{Formulations}
\label{sec:pevlm_eq}
The computational formulation of PEVLM is defined as:
\begin{align}
    &\text{Attn}s = f(Q_s, K_s, V_s), \\
    &\text{Attn}_{c_i} = f(Q_{c_i}, K_{s+c_0+\dots+c_{i-1}}, V_{s+c_0+\dots+c_{i-1}}), \\
    &\text{Attn}_q\ = f(Q_q, K_{s+c_{\text{all}}+q}, V_{s+c_{\text{all}}+q}),
\end{align}
where $s$ denotes the Sink Block, $q$ denotes the Question Block, and $c_i$ denotes the $i$-th Context Block. Symbols $Attn_s$, $Attn_{c_{i}}$, and $Attn_q$ represent the attention outputs for the corresponding blocks. The function $f(\cdot)$ corresponds to the standard softmax attention defined in Equation~(\ref{eq:fullattn}), with $Q$, $K$ and $V$ denoting the query, key, and value matrices respectively.

The total computational operation (OP) count of PEVLM is:
\begin{align}
    &\text{OP}_{\text{PEVLM}}=\text{OP}_{\text{Sink}}+N\times \text{OP}_{\text{ContextBlock}}+\text{OP}_{\text{Quest}}, \\
    &\text{OP}_{\text{Sink}}=2HS^2, \label{eq:attn_sink} \\
    &\text{OP}_{\text{ContextBlock}}=2HB(S+B), \label{eq:attn_blocks} \\
    &\text{OP}_{\text{Quest}}=2HQL, \label{eq:attn_quest}
\end{align}
where $S$ denotes the sink block size, $N$ denotes the context block number, $B$ denotes the context block size, $Q$ denotes the query block size and $H$ is the hidden size. $L$ is the total token number and $L=S+N\times B+Q$.

\noindent After simplification:
\begin{align}
\text{OP}_{\text{PEVLM}}=2H(&S^2+Q^2+NB^2 \nonumber \\
                     &+QS+NQB+NSB).
\label{eq:attn_PEVLM}
\end{align}
As the context block size is a static number, the computational complexity of PEVLM is $O(L)$.

\subsubsection{Block Size}
Since the information within each video frame represents spatial content at a specific moment, while information across frames reflects temporal dynamics, videos naturally possess structural boundaries. If the video are divided into sink and context blocks by token count, this may compromise the spatial integrity of the boundary frames shared between adjacent blocks. So we divide the video by frame.

As analyzed in the previous section, the number of frames selected for the Sink Block depends primarily on the distribution of attention weights. The optimal strategy is to include all tokens with significantly higher attention scores at the beginning of the video in the Sink Block. However, on the one hand, the number of frames containing visual tokens with a high attention score varies across different videos; on the other hand, as shown in Equation~\ref{eq:attn_PEVLM}, a larger sink block size leads to higher latency, resulting in a trade-off between accuracy and efficiency. A similar trade-off also exists in determining the context block size. We will further analyze this in the following sections.

\section{Experiments}
\label{sec:Experiments}

In this section, we test PEVLM from multiple perspectives against several different VLMs. The primary objective is to evaluate the accuracy, computational efficiency, and real-world applicability of PEVLM for long-video processing.
And to better reflect practical deployment scenarios, we designed the experiments from two perspectives.
(i) We first test with workloads of approximately 100k tokens to assess the acceleration performance of PEVLM in cloud serving platforms.
(ii) Furthermore, we evaluate PEVLM on the LongVideoBench dataset under a given latency constraint, in order to measure its effectiveness in edge scenarios where both computational resources and latency are limited.

\subsection{Accuracy Evaluation}
\label{sec:accuracy}
This experiment aims to evaluate the impact of PEVLM on accuracy in long-video understanding tasks. 
\subsubsection{Setup}
We adopt LongVideoBench \citep{longvideobench2024}, VideoMME \citep{fu2024video}, EgoSchema \citep{NEURIPS2023_90ce332a}, and MVBench \citep{li2024mvbenchcomprehensivemultimodalvideo} as benchmarks, which are designed to comprehensively evaluate multimodal models in video understanding. All evaluations are conducted using the lmms-eval toolkit \citep{lmms_eval2024}. We test on LLaVA-Video \citep{zhang2024videoinstructiontuningsynthetic}, LongVILA \citep{chen2024longvilascalinglongcontextvisual}, and Qwen2.5-VL \citep{bai2025qwen25vltechnicalreport} as representative models. To minimize accuracy loss, all experiments are conducted with bf16 precision.

We compared the accuracy of Full-Attention, Block-Attention, APE, Star-Attention, and PEVLM in the datasets. Since the other methods are finetune-free, we did not finetune the weights in the Block Attention test. For APE, we conducted experiments with a temperature setting of $T=1.0$. This choice was motivated by two considerations. First, lowering the temperature further leads to additional accuracy degradation. As previously observed, parallel encoding tends to produce lower attention weight distributions over context blocks in VLMs, rather than higher. Therefore, decreasing the temperature further skews the attention distribution away from that of Full-Attention, exacerbating the accuracy loss. Second, using $T=1.0$ is favorable for large-scale deployment. Regarding the sink block (shared prefix) size, we selected the system prompt by referring to the code provided by APE. For Star-Attention, we implemented its equivalent algorithm on a single node, with the anchor size set equal to the context block size, following the setup in the Block-Attention paper \citep{ma2025blockattentionefficientprefilling}. All methods use a context block size of 4096 tokens. Smaller context block sizes resulted in reduced accuracy for all methods, while larger blocks significantly degraded inference performance. For PEVLM, we set the sink and context block size to 16 frames, which corresponds to approximately 4k tokens for Qwen2.5-VL and LongVILA, and about 3k tokens for LLaVA-Video.

\subsubsection{Results}
As shown in Table~\ref{tab:algo_acc}, for shorter video scenarios, all methods achieve accuracy comparable to or even better than Full-Attention.
Although parallel encoding is generally expected to degrade performance, it can actually improve accuracy in long-context scenarios. This is because the softmax function in attention mechanisms struggles to produce reliable outputs when the input length exceeds the training horizon \citep{2025softmaxforsharpsize}. However, parallel encoding effectively reduces the number of tokens involved in each softmax computation, thus alleviating the degradation of softmax behavior in long-context settings. As a result, the softmax output more robustly approximates an ideal sharp distribution, ultimately enhancing the performance of certain models under extended context lengths.

While all parallel encoding methods perform well on short videos, their effectiveness declines as the temporal length of the video increases, especially when video durations extend to several or even tens of minutes. In such scenarios, PEVLM consistently outperforms other parallel encoding approaches, particularly on the Qwen2.5-VL model.
Qwen2.5-VL uses 3D-MROPE and is thus more sensitive to temporal order in videos. As APE and Star-Attention reuse position IDs across different context blocks, which severely distort temporal features, resulting in meaningless or garbled output. In contrast, Block-Attention reencodes positional information, which improves its accuracy on Qwen2.5-VL. PEVLM further improves accuracy by more rigorously preserving sequential positional encodings and incorporating a shared sink block, leading to better results.
For LongVILA and LLaVA-Video, which use standard RoPE, the aforementioned issues are less pronounced. Nonetheless, PEVLM still outperforms other parallel encoding methods in long-video scenarios.

\subsection{Performance Evaluation}
In this section, we evaluate the performance of PEVLM.
{\setlength{\intextsep}{2pt}
\begin{figure}[H]
    \includegraphics[width=0.47\textwidth]{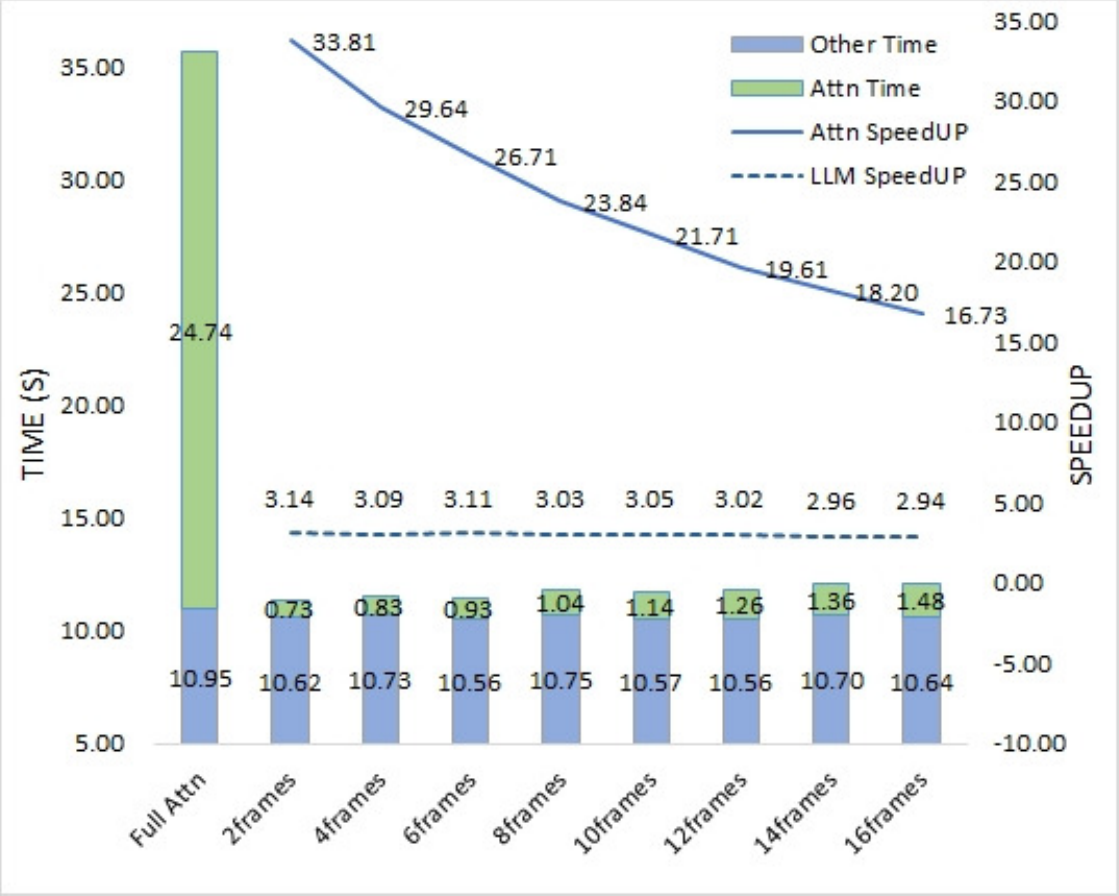}
    \caption{PEVLM under different context\_block\_size settings. "Attn Time" refers to attention computation time, and "Other Time" covers all other LLM costs. "Attn SpeedUP" and "LLM SpeedUP" indicate the acceleration over Attention layer and overall LLM performance. The sink\_size is set to (system prompt+one frame).}
    \label{fig:perf_block}
\end{figure}

\begin{figure}[H]
    \includegraphics[width=0.47\textwidth]{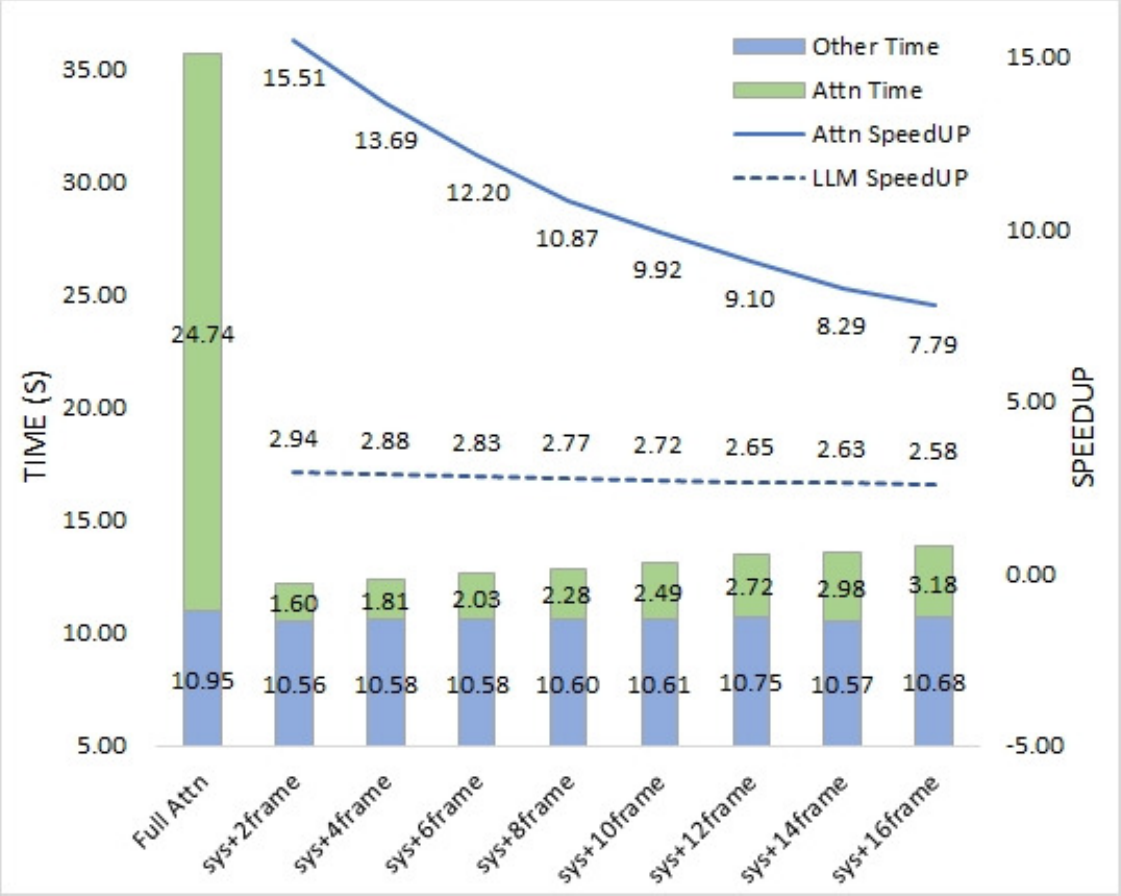}
    \caption{PEVLM under different sink\_size settings. "Attn Time" refers to attention computation time, and "Other Time" covers all other LLM costs. "Attn SpeedUP" and "LLM SpeedUP" indicate the acceleration over Attention layer and overall LLM performance. The context\_block\_size is set to (16 frames).}
    \label{fig:perf_sink}
\end{figure}
}
\subsubsection{Setup}
To facilitate experimentation, we built PEVLM based on the SGLang, which is a fast serving framework for large language models and vision-language models, widely used in cloud production deployment.
We conducted computational efficiency evaluations on the Qwen2.5-VL model. The primary objective of this experiment is to assess the improvements in computational efficiency introduced by PEVLM. All experiments were carried out on an NVIDIA H20-96G GPGPU. Given that PEVLM primarily optimizes the attention mechanism within the LLM, we measured the execution time of both the entire LLM module and its individual attention layers separately.

\subsubsection{Results}
As shown in Figure~\ref{fig:perf_block} and Figure~\ref{fig:perf_sink}, the end-to-end inference speed of the LLM improves by 2.58× to 3.14× as the block size and sink size vary. The attention module, in particular, achieves substantial acceleration, with speedups ranging from 7.79× to 33.81×, while the runtime of non-attention components remains largely stable. As the size of the sink block or context blocks increases, the speedup of the attention module correspondingly decreases. This trend is consistent with Equation~\ref{eq:attn_PEVLM}, which indicates that the computational cost of attention grows quadratically with the size of the sink and context blocks. However, since non-attention components account for a considerable portion of total runtime, the overall end-to-end speedup degrades only moderately. However, as the context blocks size and sink block size increase to 16 frames, the proportion of attention in the total runtime increases from 6.4\% to 22.9\%, indicating that further increases in block size will lead to a more pronounced drop in overall performance. We will analyze the trade-off between performance degradation and potential accuracy gains from larger sink/context blocks size in the following sections.

\subsection{Accuracy with Limited Latency}
\label{sec:latency_limit}
Deploying VLMs on edge devices presents significant challenges due to constrained computational resources and stringent latency requirements. To evaluate the practical benefits of PEVLM in such settings, we designed a latency-aware simulation that mimics real-world deployment conditions on resource-limited devices.

\subsubsection{Setup}
We adopt LongVideoBench as the evaluation benchmark and introduce a latency threshold during inference: any sample that exceeds this latency limit is treated as a failure. All experiments are conducted on the  NVIDIA H20-96G GPGPU.
We use Qwen2.5-VL-7B-Instruct with Full-Attention as the primary baseline, and compare the performance of Full-Attention and PEVLM under identical latency constraints.
To further reflect deployment in low-resource environments, where smaller models are commonly used to meet latency budgets, we also include comparisons with Qwen2.5-VL-3B-Instruct, a smaller variant Qwen2.5-VL-7B-Instruct.

\begin{table}[ht]
    \centering
    \caption{Accuracy with Limited Latency}
    \label{tab:lantency_acc}
    \begin{tabular}{l l l l l l l}
        \toprule
        & Full Attn & Full Attn & \multicolumn{2}{c}{PEVLM} \\
        &  & (3B) &  \\
        \arrayrulecolor{gray!80}\midrule
        \arrayrulecolor{black}
        sink & - & - & sys+2f & sys+16f \\
        \arrayrulecolor{gray!80}\midrule
        \arrayrulecolor{black}
        context & - & - & 16f & 16f \\
        \midrule
        No Limit  & 60.43\% & 55.05\% & 61.18\% & \textbf{62.23\%} \\
        40s  & 59.69\% & 55.05\% & 61.18\% & \textbf{62.23\%} \\
        30s  & 27.08\% & 55.05\% & 61.18\% & \textbf{62.23\%} \\
        20s  & 23.26\% & 24.38\% & \textbf{61.03\%} & 60.28\% \\
        \bottomrule
    \end{tabular}
\end{table}

\subsubsection{Results}
As shown in Table~\ref{tab:lantency_acc}, the accuracy of Full-Attention (7B) drops sharply when the latency constraint is reduced from 40s to 30s, and the smaller Full-Attention (3B) model also shows a significant decline when the limit is further tightened to 20s.
In contrast, PEVLM maintains consistently high accuracy even under the strictest 20s constraint. Notably, while the larger sink size (sys+16f) achieves the best accuracy under relaxed constraints ($\ge$30s), the smaller configuration (sys+2f) performs better under the 20s limit due to its lower latency overhead.

In general, PEVLM significantly improves accuracy in latency-constrained scenarios. Moreover, to maximize performance across diverse hardware and deployment conditions, the optimal configuration (e.g., sink/block size) should be adaptively tuned.

\section{How does each component in PEVLM contribute to the performance?}
\label{sec:components}
\begin{table}[h]
    \centering
    \caption{Ablation study of PEVLM components on LongVideoBench}
    \label{tab:components}
    \begin{tabular}{c c c|c c}
        \toprule
         SP & DF & FS & LongVideoBench & VideoMME \\
        \midrule
         & & & 46.67\% & 45.31\% \\
         $\checkmark$ & & & 56.75\% & 60.51\%  \\
         $\checkmark$ & $\checkmark$ & & 56.77\% & 60.63\% \\
         $\checkmark$ & $\checkmark$ & $\checkmark$ & 58.26\% & 62.22\%  \\
        \bottomrule
    \end{tabular}
\end{table}

In Table~\ref{tab:components}, we present an ablation study to evaluate the contribution of each component in PEVLM, including the use of Sequential Position Embedding (SP), Dividing the Video by Frames (DF), and Adding Frames to the Sink (FS). The results are averaged across the four base models reported in Table~\ref{tab:algo_acc}. We observe that sequential parallel encoding yields the most significant improvement in accuracy, achieving an average performance gain of 12.64\%. Incorporating the initial frames into the sink block leads to an additional average accuracy gain of 1.54\%. In contrast, dividing the video by frames provides only a marginal improvement compared to directly segmenting the video without this strategy.

\subsection{Sequential Position Embeddin}
We further analyze the component with the greatest impact on accuracy: the preservation of sequential position embeddings. As discussed before, adopting sequential position embeddings instead of reusing position embeddings across context blocks produces attention distributions that more closely resemble those of Full-Attention, thereby aligning better with the model's expected behavior. As shown in Table~\ref{tab:pos_acc}, the use of sequential position embeddings significantly improves accuracy in both Qwen2.5-VL and LongVILA. In contrast, for the LLaVA-Video model, the benefit is marginal, likely because its attention weight distribution (as shown in Figure~\ref{fig:llava_attention_weights} in the Appendix) remains relatively stable even when using parallel encoding.

\begin{table}[h]
    \centering
    \caption{Accuracy $\uparrow$ on LongVideoBench dataset with Different Position Embedding Strategy}
    \label{tab:pos_acc}
    \begin{tabular}{l l l l l}
        \toprule
        model & method & reuse pos & sequential \\
        \midrule
         Qwen2.5-VL- & APE (T=1.0) & 30.44\% & 59.39\% \\
         7B-Instruct & Star Attn & 29.54\%  & 61.41\% \\
         & PEVLM & 14.73\% & 62.23\% \\
        \midrule
         LongVILA- & APE (T=1.0) & 51.46\% & 53.40\% \\
         7B-256f & Star Attn & 51.76\% & 53.63\% \\
         & PEVLM & 49.96\% & 53.70\% \\
        \midrule
         LLaVA-Video- & APE (T=1.0) & 58.19\% & 58.12\% \\
         7B-Qwen2 & Star Attn & 58.71\% & 58.49\% \\
         & PEVLM & 58.34\% & 59.01\% \\
        \bottomrule
    \end{tabular}
\end{table}


\subsection{Sink \& Context Block Size}
\label{sec:sink_size}

As analyzed in the \textit{PEVLM} section, the inference latency of PEVLM grows quadratically with the sizes of both the sink and context blocks, which aligns with the empirical results shown in Figure~\ref{fig:perf_block} and Figure~\ref{fig:perf_sink}. We further investigate how these sizes affect model accuracy.

\begin{figure}[h]
    \centering
    \includegraphics[width=0.47\textwidth]{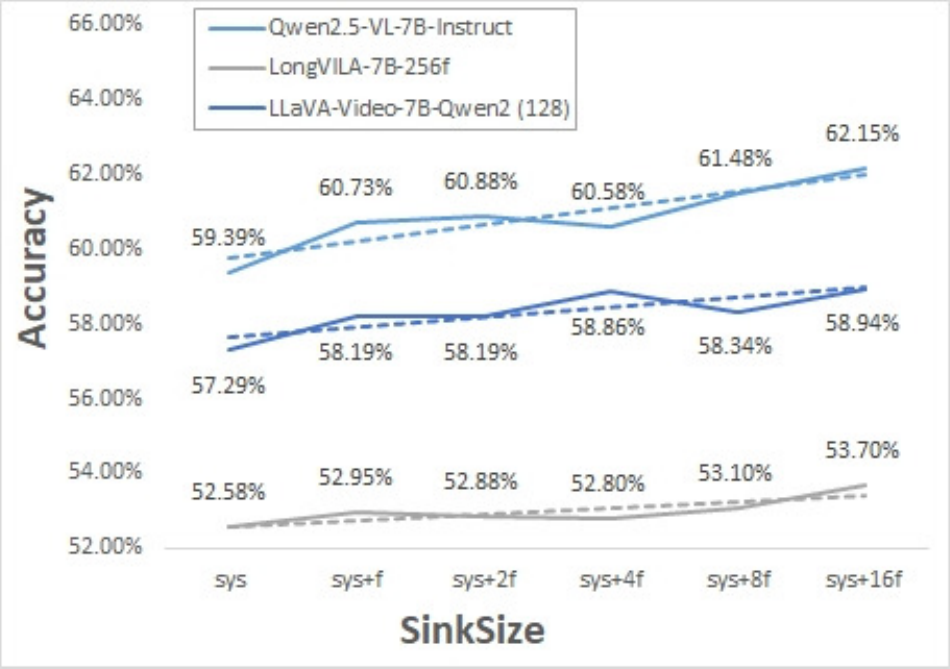}
    \caption{The impact of sink sizes on accuracy}
    \label{fig:sink_size}
\end{figure}

As noted in the \textit{Observations} section, including the initial frames of the video in the sink is crucial. This is validated in Figure~\ref{fig:sink_size}, where incorporating early frames into the sink leads to noticeable accuracy improvements across all evaluated models.

Figure~\ref{fig:sink_size} also shows that model accuracy generally increases with sink size. However, due to the quadratic latency growth, excessively large sink sizes are not ideal for deployment. A simple pre-deployment test may be necessary to determine the optimal configuration. For the three models in Figure~\ref{fig:sink_size}, a sink size between 14 and 18 already achieves strong performance. When latency constraints are considered, this leads to a trade-off between accuracy and efficiency. For example, as shown in Table~\ref{tab:lantency_acc}, under a 20-second latency budget, using only \texttt{[SYS] + 2 frames} as the Sink Block yields higher accuracy than \texttt{[SYS] + 16 frames}.

\begin{figure}[h]
    \centering
    \includegraphics[width=0.47\textwidth]{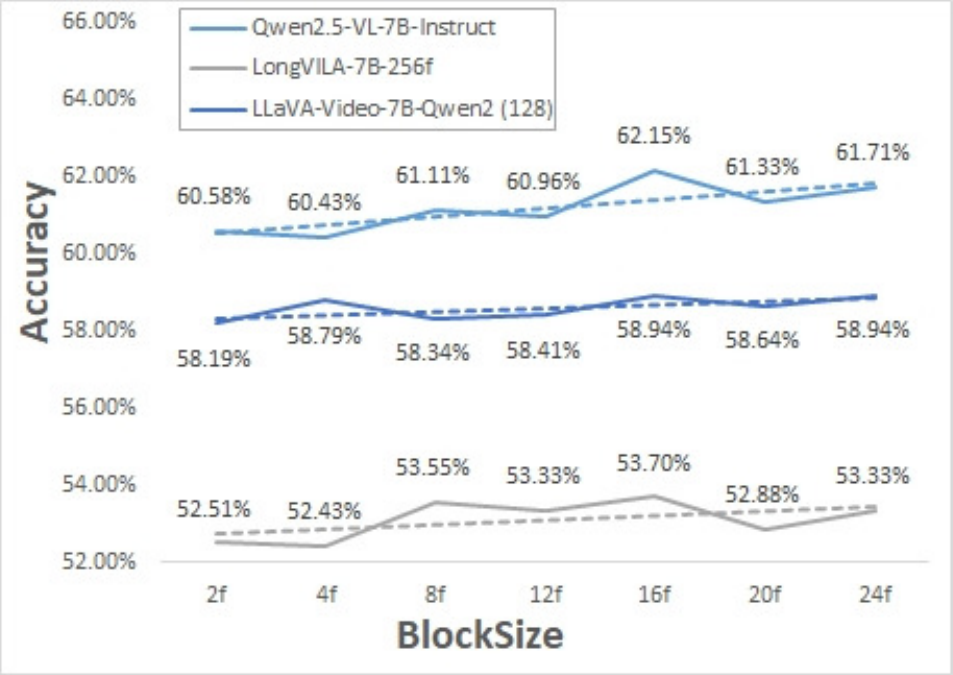}
    \caption{The impact of block sizes on accuracy}
    \label{fig:block_size}
\end{figure}

We also examine the effect of context block size on model accuracy. As shown in Figure~\ref{fig:block_size}, increasing the size of context blocks leads to consistent accuracy gains, similar to the sink size. However, Figure~\ref{fig:perf_block} reveals that latency also increases quadratically with block size, highlighting the same trade-off between accuracy and computational efficiency.

\section{Conclusion}
We introduce PEVLM, a fine-tuning-free parallel encoding strategy that significantly accelerates Vision-Language Models for long-video understanding. By preserving sequential position embeddings and leveraging a shared Sink Block, PEVLM aligns attention behavior with Full-Attention while reducing complexity from $O((T\times N)^2)$ to $O(T\times N)$. It achieves up to 7.47$\times$ attention speedup, 40\% end-to-end latency reduction, and strong accuracy gains under tight latency constraints. PEVLM offers a practical and scalable solution for real-world, long-context multimodal tasks.

\section{Future Work}

Although PEVLM has demonstrated substantial improvements in computational efficiency and reasoning performance for long video processing tasks, there remains ample opportunity for further research and enhancement. In future work, we plan to explore the following directions.

Beyond video inputs, an important future direction is extending PEVLM to handle richer multimodal inputs such as LiDAR, map priors, and inertial measurements, which are commonly used in domains like autonomous driving and robotics. These modalities often exhibit different spatial and temporal characteristics compared to visual tokens, posing unique alignment and fusion challenges for attention-based models. Enhancing PEVLM with cross-modal position encoding strategies and adaptive sink block selection could further improve its generalization to complex, heterogeneous input streams.

One promising direction is extending PEVLM to streaming or online scenarios, such as autonomous driving, where the model continuously receives multimodal inputs (e.g., video frames and LiDAR) over time. In these settings, a fixed-length sliding temporal window is often adopted: new data is appended while the oldest is discarded, maintaining a constant context size that reflects the most recent few seconds or minutes of input. If PEVLM can be adapted to such settings, it would allow the model to compute only over the newly arrived data at each inference step, rather than reprocessing the entire history. This could lead to significant performance gains—potentially improving inference speed by several times or even orders of magnitude. Such an extension would enable real-time video-language reasoning under strict latency constraints and dynamic, continuously evolving environments.

One additional direction for future exploration is to investigate why parallel encoding methods can surpass Full-Attention in accuracy. Our preliminary analysis suggests that this may be due to the shortened input length in the attention softmax, which improves attention sharpness and reliability in long-context settings. However, since this work focuses on applying and optimizing parallel encoding for VLMs, we leave a deeper theoretical and empirical investigation of this phenomenon for future work. Understanding its root cause may offer new insights into further boosting model performance on long contexts.


\FloatBarrier
\bibliography{aaai2026}

\newpage

\onecolumn

\appendix

\section*{Appendix A. Benchmark Details}
We evaluate our method on several video understanding benchmarks that test different aspects of video comprehension:

\subsection{EgoSchema}
EgoSchema \citep{NEURIPS2023_90ce332a} is a large-scale benchmark designed to evaluate Multimodal Large Language Models (MLLMs) on egocentric video understanding. It consists of 100 hours of first-person video data spanning 1,270 daily activity episodes across diverse real-world environments. The benchmark introduces over 10,000 manually curated question-answer pairs, covering tasks such as object grounding, human-object interaction, activity reasoning, and intent prediction.

\subsection{MVBench}
MVBench \citep{li2024mvbenchcomprehensivemultimodalvideo} is a comprehensive benchmark designed to evaluate Multimodal Large Language Models (MLLMs) on multi-granular video understanding. It consists of 5 task categories—including moment-level, frame-level, clip-level, video-level, and holistic video understanding—covering a wide range of temporal scopes and reasoning demands. The benchmark includes 2,562 manually annotated questions grounded in 4,119 diverse video clips, selected from real-world scenarios.

Each question is carefully designed to probe different levels of spatiotemporal understanding, from fine-grained object recognition and short-term motion tracking to long-term event inference. MVBench offers a unified and challenging evaluation protocol to assess the generalization and reasoning ability of MLLMs across granularities.

Unlike conventional third-person video datasets, EgoSchema emphasizes embodied perception and temporal reasoning from an egocentric perspective, posing unique challenges for spatial understanding, long-term memory, and causal inference in MLLMs.

\subsection{Video-MME}
Video-MME \citep{fu2024video} is a comprehensive evaluation benchmark for assessing the video understanding capabilities of Multimodal Large Language Models (MLLMs). It spans 6 primary visual domains and 30 subfields, covering a diverse range of video types and temporal scenarios. The benchmark includes 900 videos with durations ranging from 11 seconds to 1 hour, totaling 254 hours of content.

To evaluate fine-grained visual and temporal reasoning, 2,700 manually annotated question-answer pairs are provided. Video-MME challenges models to comprehend both short and long video clips across different temporal granularities, making it a rigorous benchmark for evaluating the core video processing capabilities of MLLMs.

\subsection{LongVideoBench}
LongVideoBench \citep{longvideobench2024} is a benchmark specifically designed to evaluate the long-context video understanding capabilities of Multimodal Large Language Models (MLLMs). It features 1,760 videos spanning 12 diverse real-world scenarios, with video durations ranging from 5 minutes to 2 hours, totaling over 1,000 hours of content. The benchmark includes 2,400 manually annotated multi-choice questions targeting key aspects of long video comprehension.

LongVideoBench focuses on challenging long-range temporal reasoning, event tracking, and global understanding across extended video content. It aims to assess whether models can maintain coherence, memory, and attention over prolonged contexts, making it a rigorous testbed for long-form video modeling.

\section*{Appendix B. Attention Weights Distribution of LLaVA-Video and LongVILA}
\label{app:attn_weight}

{\setlength{\intextsep}{0pt}
\begin{figure*}[ht]
    \centering
    \begin{subfigure}[t]{0.32\textwidth}
        \centering
        \includegraphics[width=\linewidth]{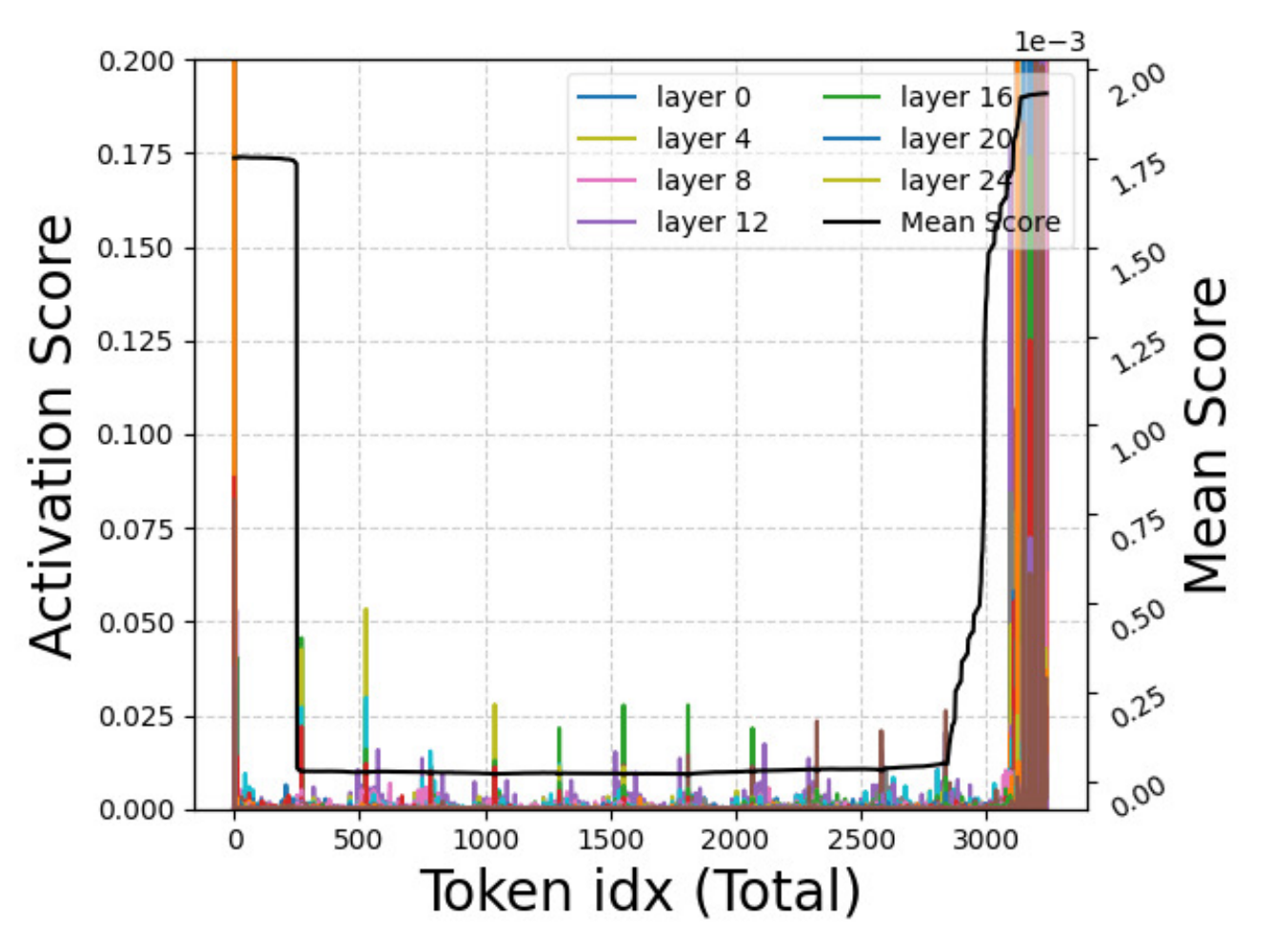}
        \includegraphics[width=\linewidth]{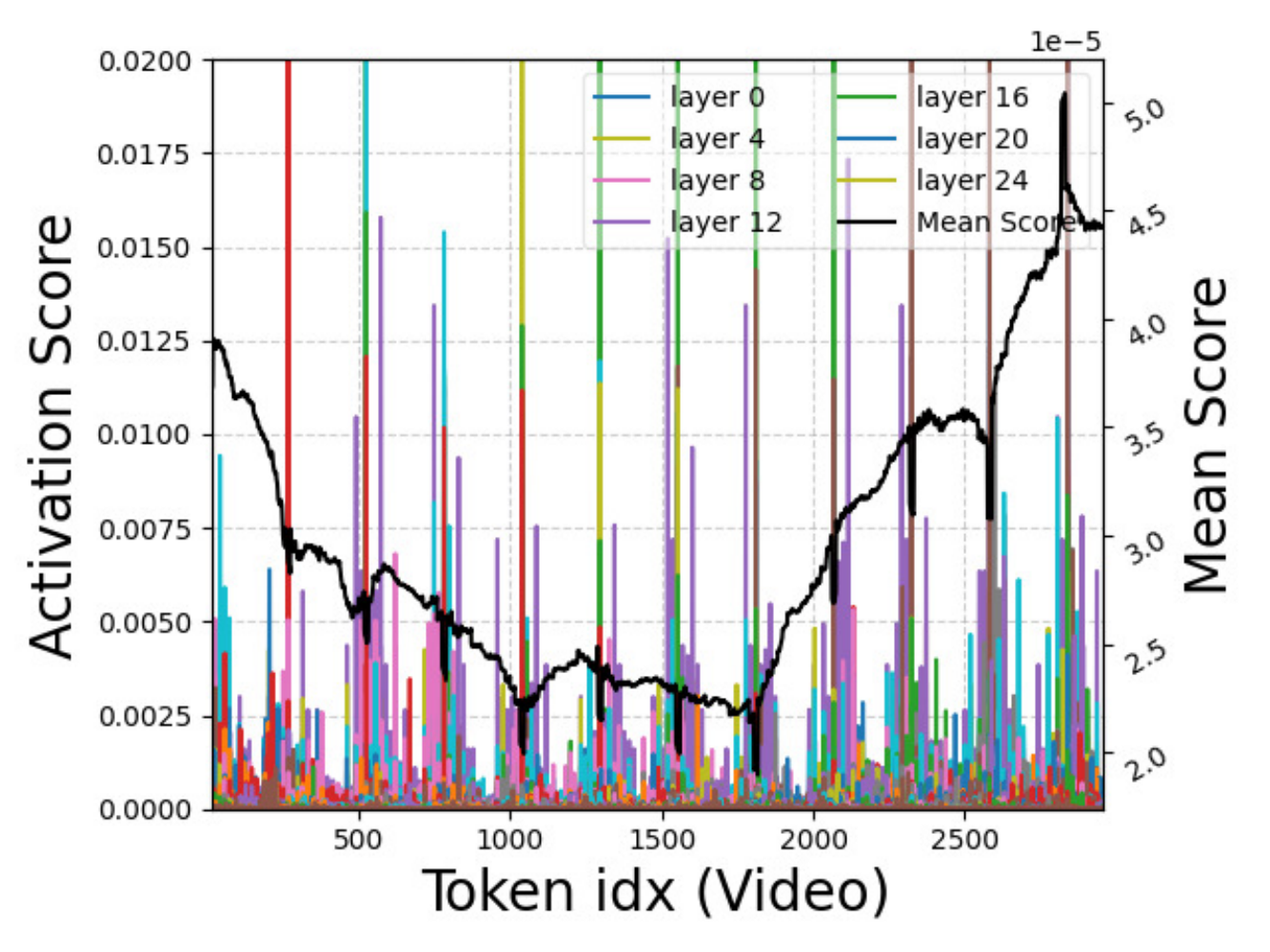}
        \caption*{(a) Full-Attention}
    \end{subfigure}
    \hfill \raisebox{-0.5\height}{\textcolor{lightgray}{\vrule width 0.8pt height 6.5cm}} \hfill 
    \begin{subfigure}[t]{0.32\textwidth}
        \centering
        \includegraphics[width=\linewidth]{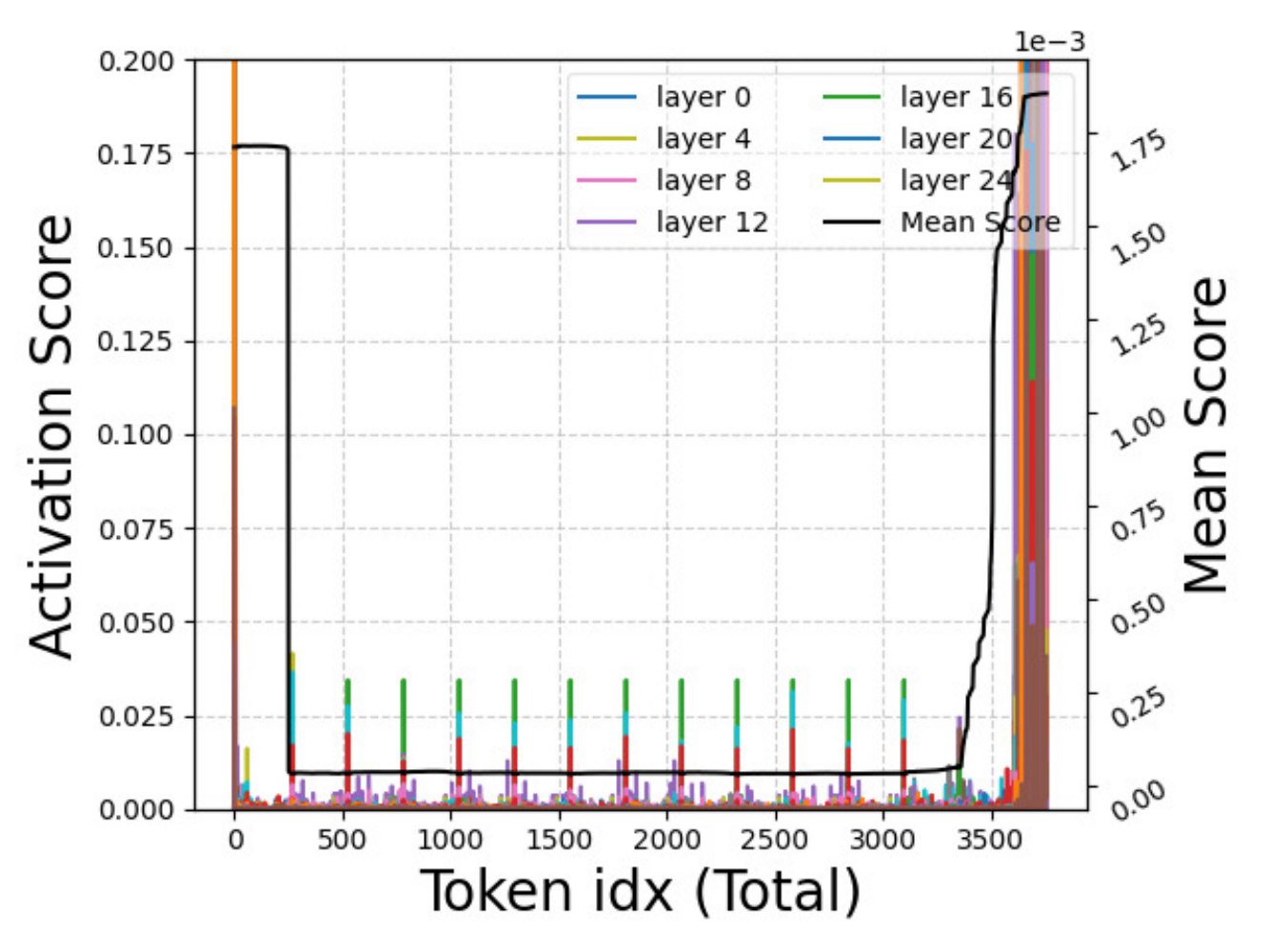}
        \includegraphics[width=\linewidth]{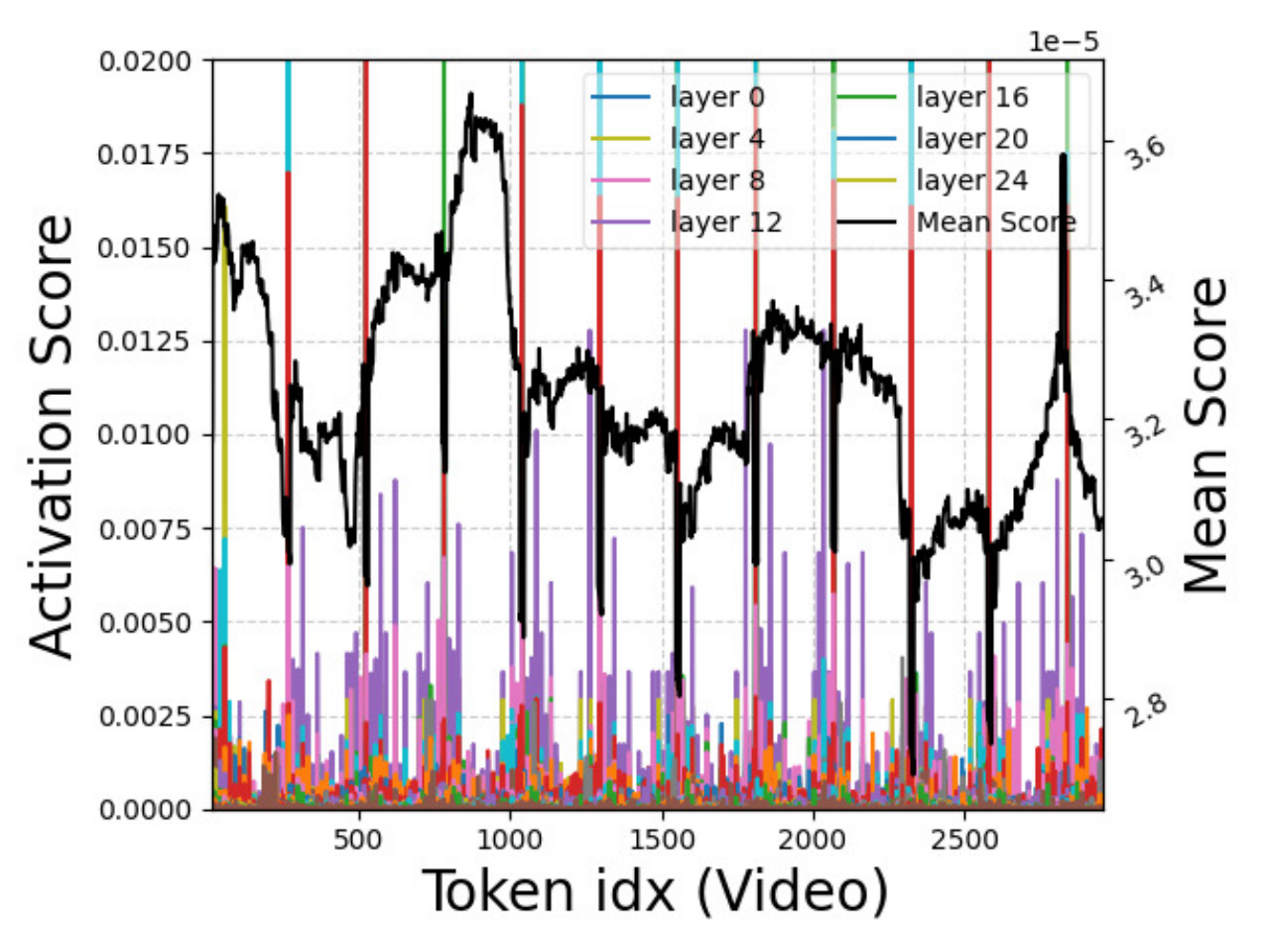}
        \caption*{(b) Reuse position embeddings}
    \end{subfigure}
    \hfill \raisebox{-0.5\height}{\textcolor{lightgray}{\vrule width 0.8pt height 6.5cm}} \hfill 
    \begin{subfigure}[t]{0.32\textwidth}
        \centering
        \includegraphics[width=\linewidth]{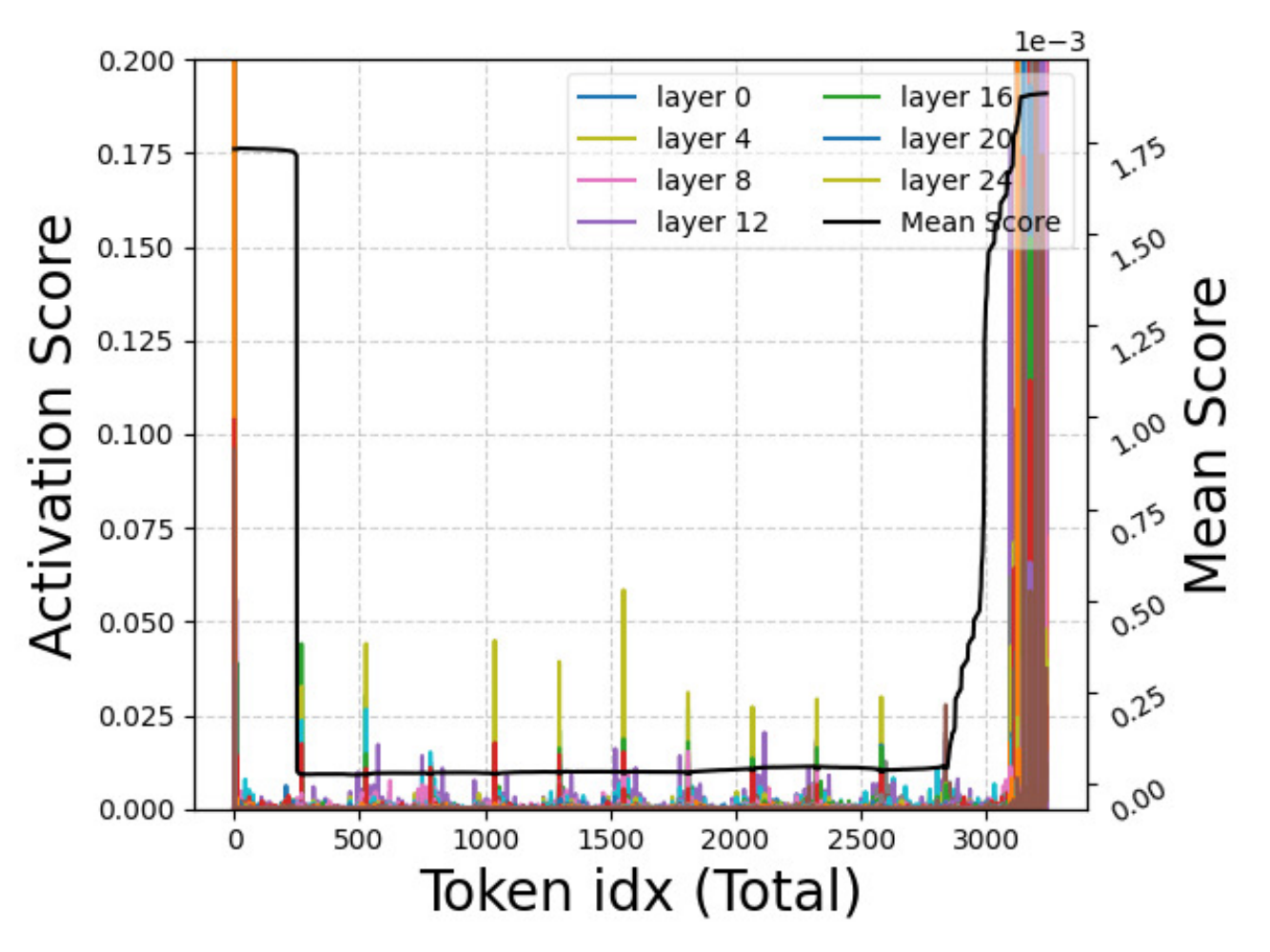}
        \includegraphics[width=\linewidth]{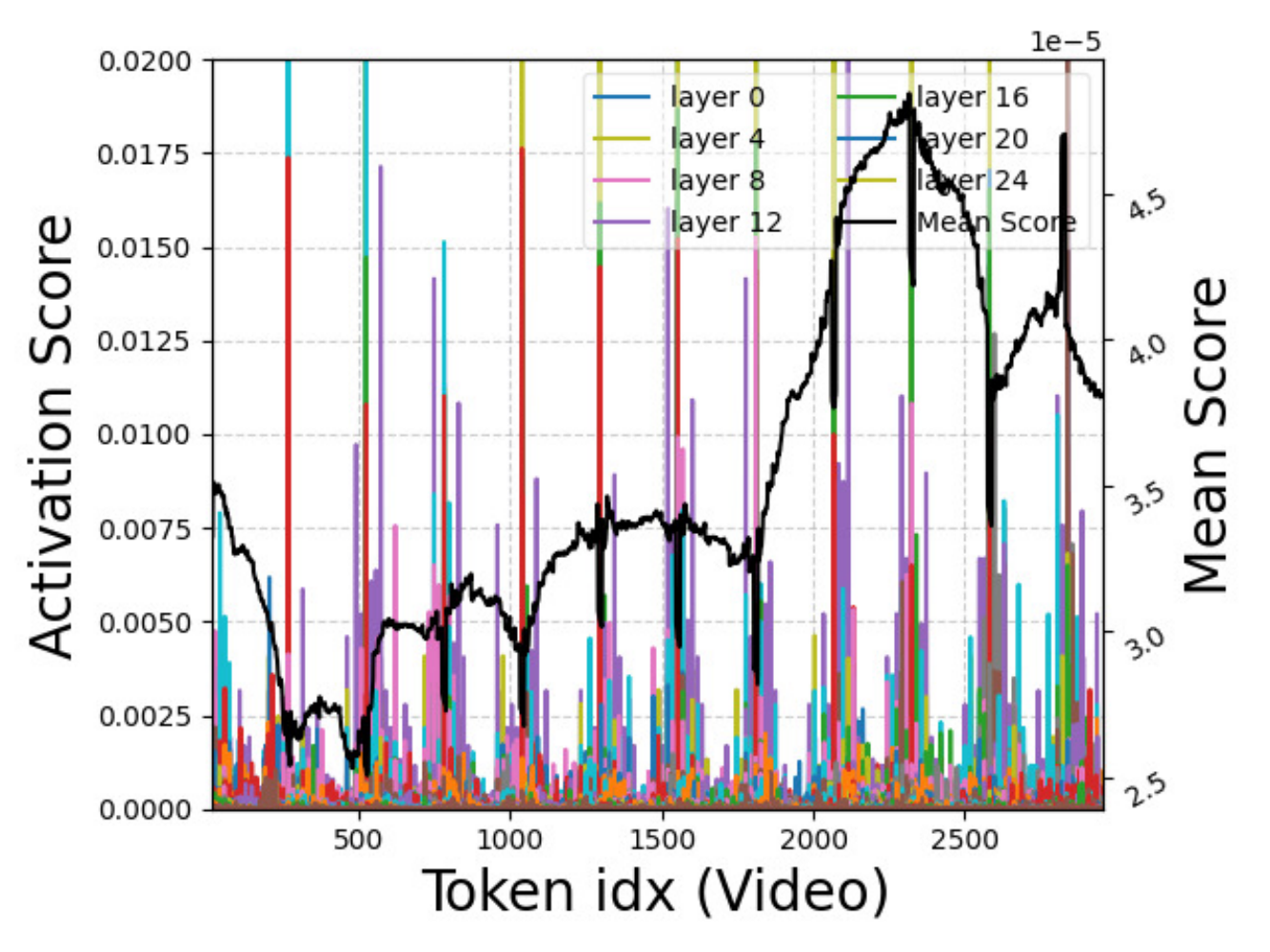}
        \caption*{(c) Sequential position embeddings}
    \end{subfigure}
    
    \caption{Attention Weight Distributions of LongVILA}
    \label{fig:vila_attention_weights}
\end{figure*}
\begin{figure*}[ht]
    \centering
    
    \begin{subfigure}[t]{0.32\textwidth}
        \centering
        \includegraphics[width=\linewidth]{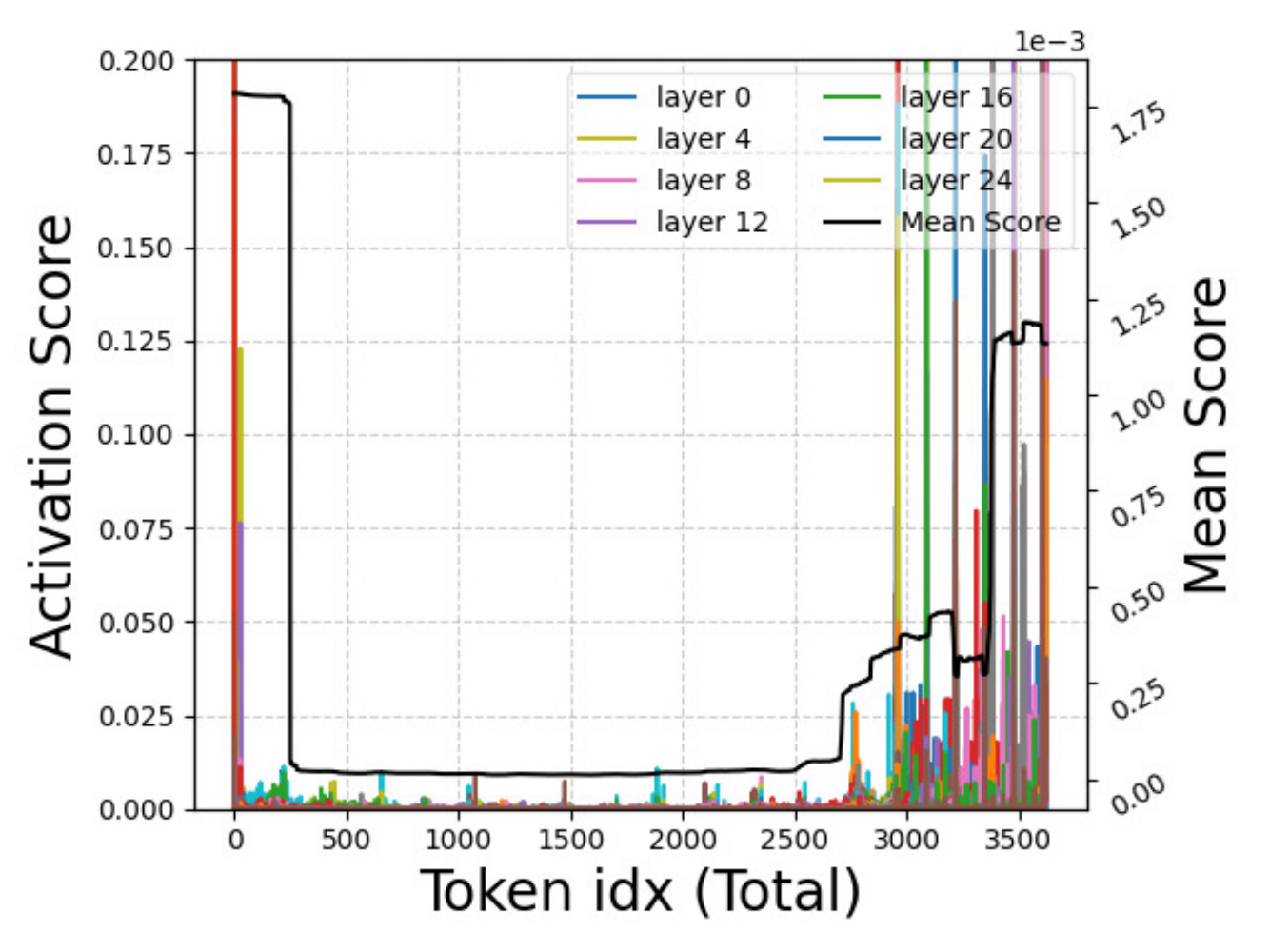}
        \includegraphics[width=\linewidth]{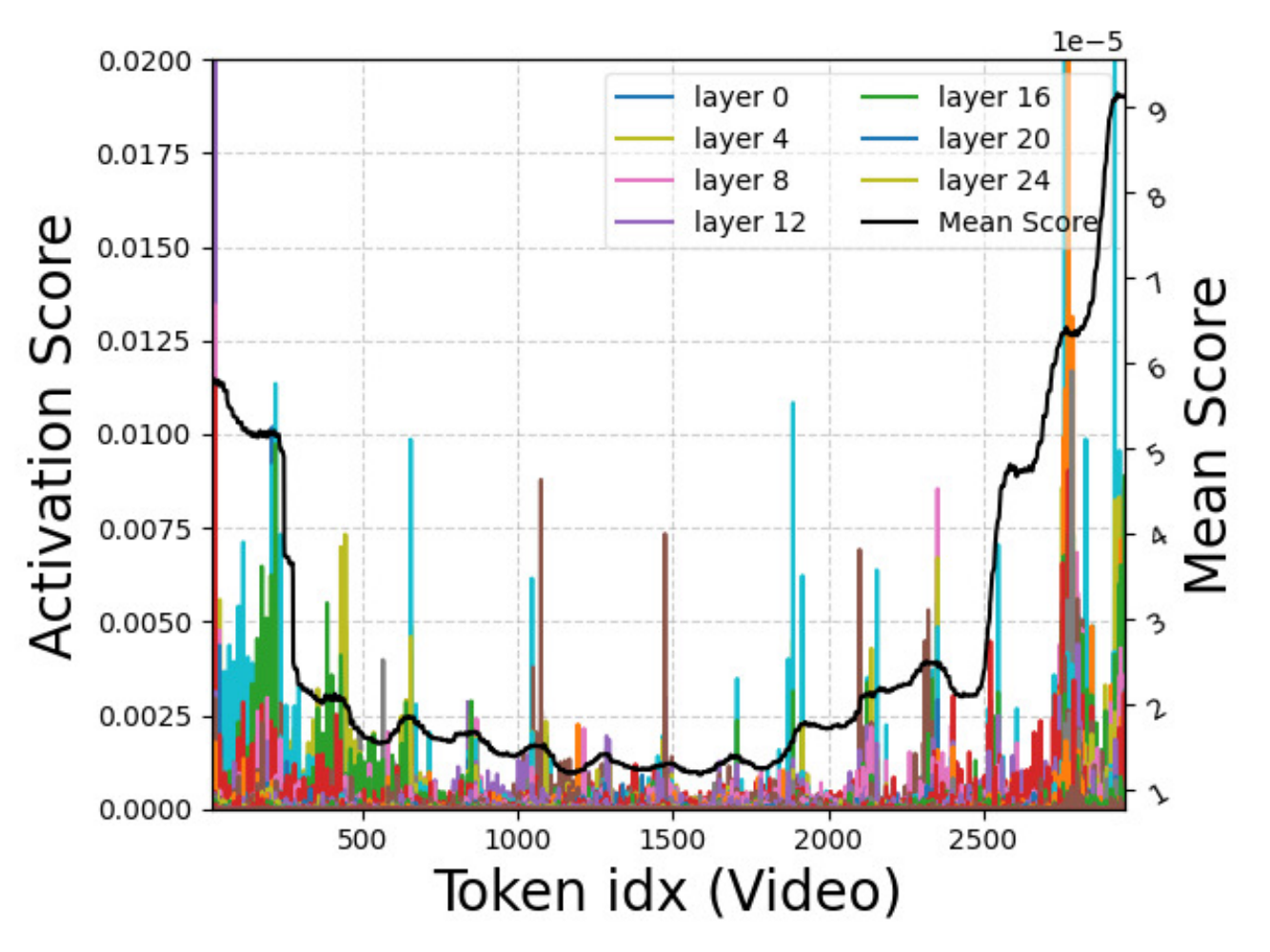}
        \caption*{(a) Full-Attention}
    \end{subfigure}
    \hfill \raisebox{-0.5\height}{\textcolor{lightgray}{\vrule width 0.8pt height 6.5cm}} \hfill 
    \begin{subfigure}[t]{0.32\textwidth}
        \centering
        \includegraphics[width=\linewidth]{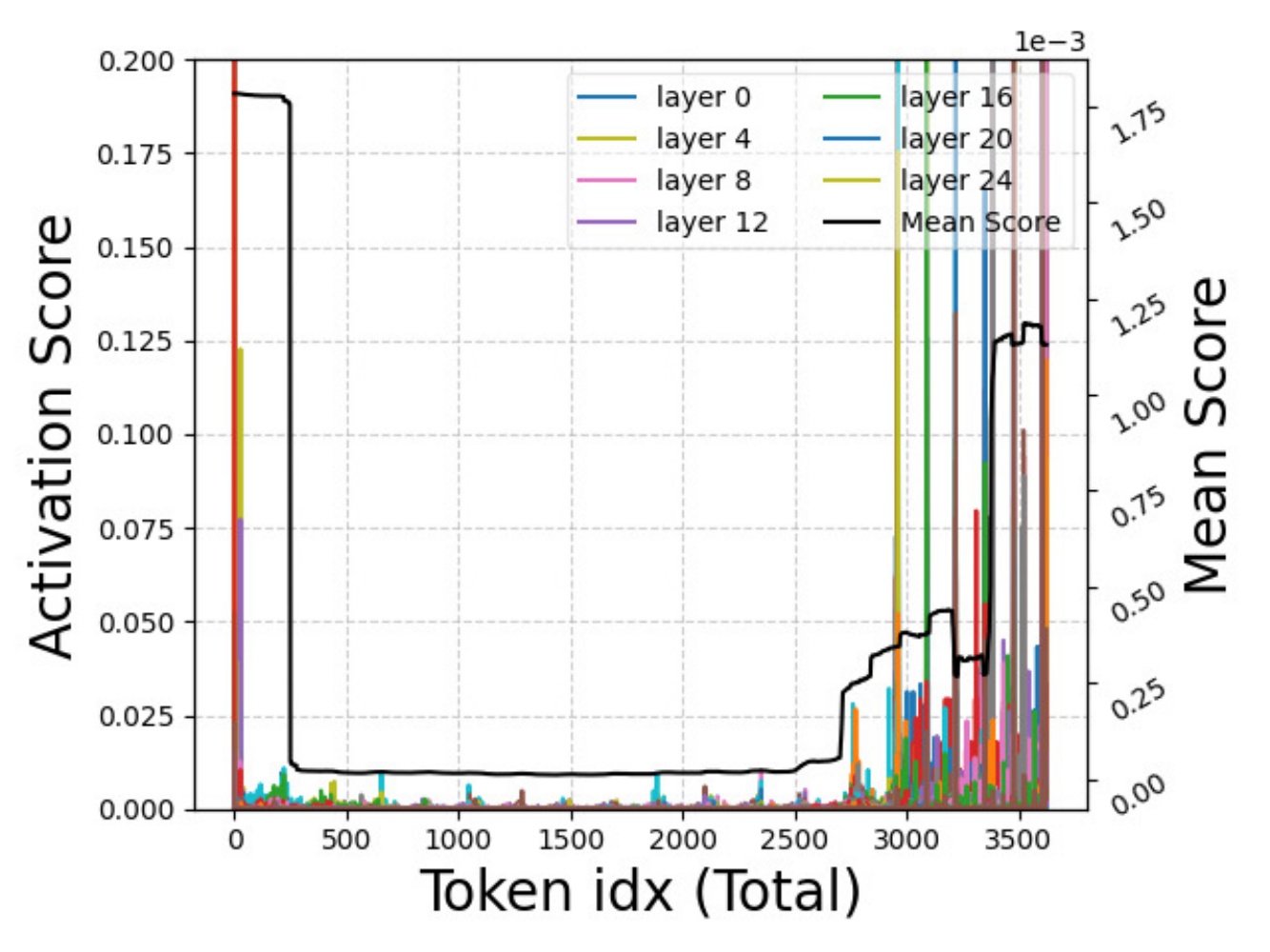}
        \includegraphics[width=\linewidth]{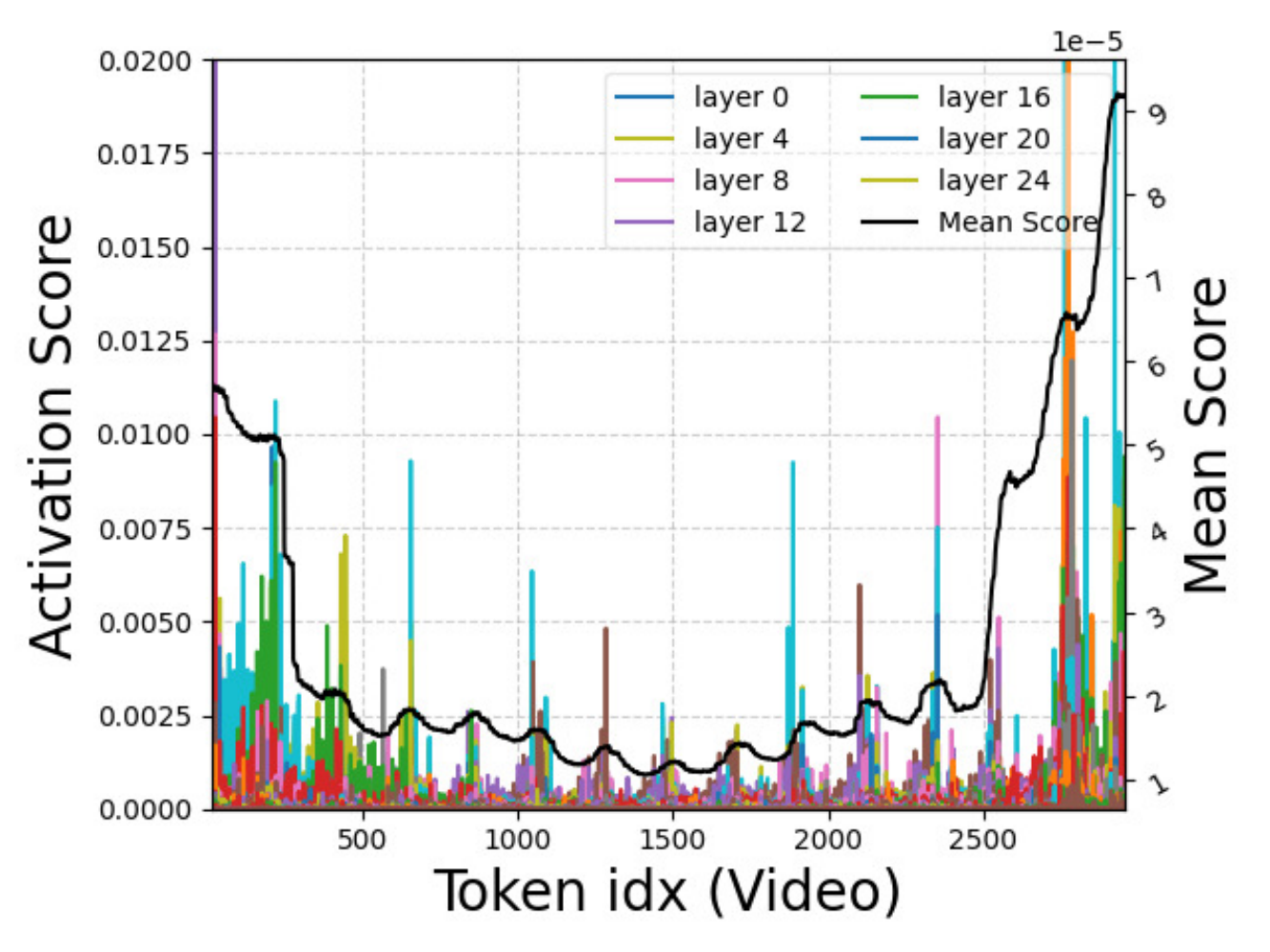}
        \caption*{(b) Reuse position embeddings}
    \end{subfigure}
    \hfill \raisebox{-0.5\height}{\textcolor{lightgray}{\vrule width 0.8pt height 6.5cm}} \hfill 
    \begin{subfigure}[t]{0.32\textwidth}
        \centering
        \includegraphics[width=\linewidth]{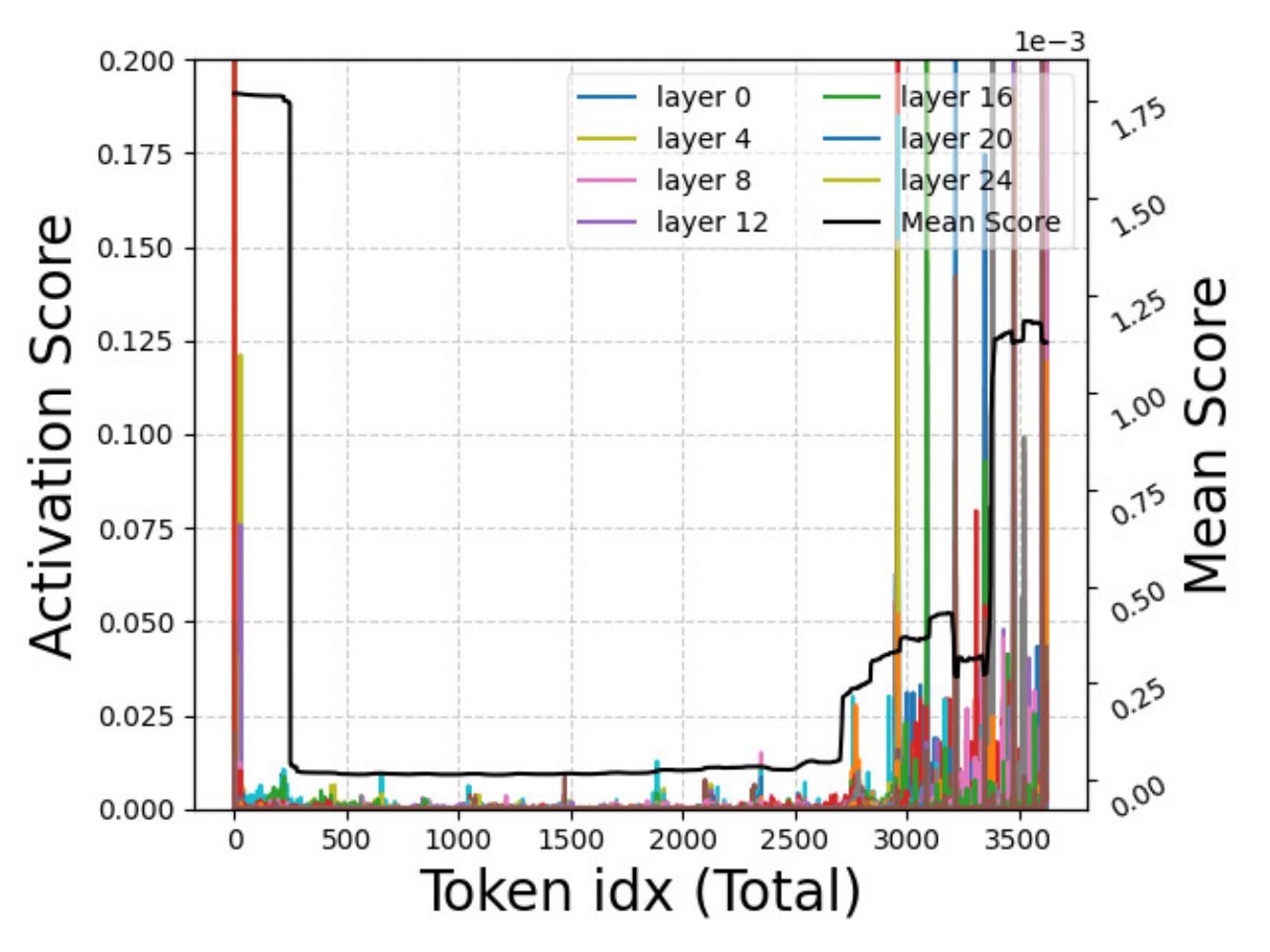}
        \includegraphics[width=\linewidth]{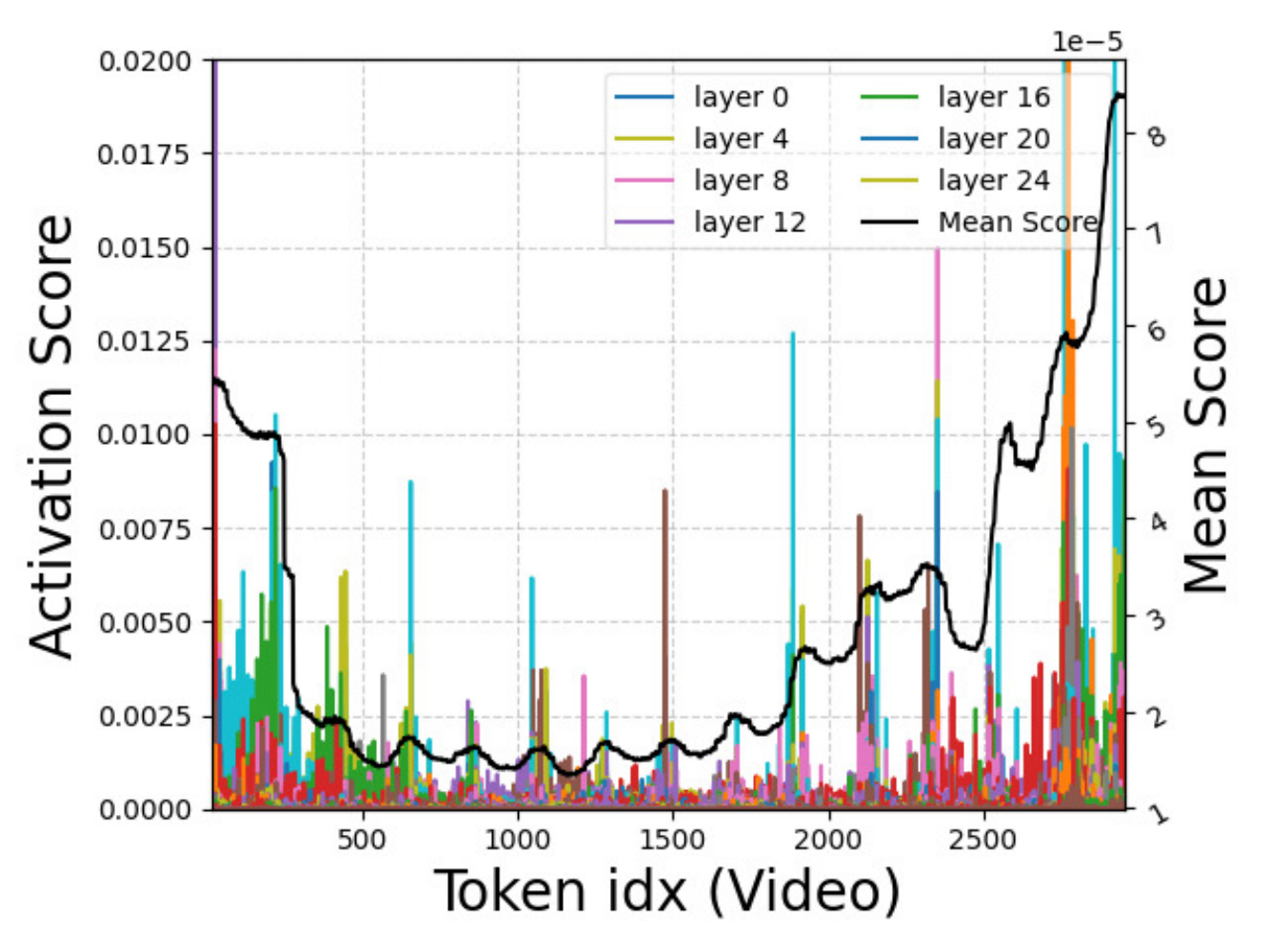}
        \caption*{(c) Sequential position embeddings}
    \end{subfigure}
    
    \caption{Attention Weight Distributions of LLaVA-Video}
    \label{fig:llava_attention_weights}
\end{figure*}
}

\end{document}